\documentclass{article}
\usepackage{iclr2027_conference,times}
\iclrfinalcopy
\pdftrailerid{}

\usepackage{amsmath}
\usepackage{booktabs}
\usepackage{tabularx}
\usepackage{graphicx}
\usepackage{capt-of}
\usepackage{wrapfig}
\usepackage{url}
\usepackage{xcolor}
\usepackage{colortbl}
\usepackage{hyperref}

\definecolor{ourblue}{rgb}{0.368,0.507,0.71}
\definecolor{ourtint}{HTML}{E6EEF8}
\definecolor{groupbg}{HTML}{FFF2CC}
\newcommand{\grouprow}[2]{\rowcolor{groupbg}[0pt][0pt]\multicolumn{#1}{@{}c@{}}{\textit{#2}\strut}}
\hypersetup{
    colorlinks,
    linkcolor=ourblue,
    citecolor=ourblue,
    urlcolor=ourblue
}

\title{Recovering Off-Policy Supervision\\ for Speculative Decoding\vspace{14pt}}

\author{%
  \vspace*{4pt}
  \makebox[5.28in][c]{%
  \begin{tabular}{@{}c@{\hspace{0.08in}}c@{}}
    \parbox[c]{2.6in}{\centering Jungseob Lee$^{1}$\\[1pt] \fontsize{8pt}{7.5pt}\texttt{omanma1928@korea.ac.kr}} &
    \parbox[c]{2.6in}{\centering Chanjun Park$^{2}$\\[1pt] \fontsize{8pt}{7.5pt}\texttt{chanjun.park@ssu.ac.kr}}\\[12pt]
    \parbox[c]{2.6in}{\centering Sugyeong Eo$^{3,\dagger}$\\[1pt] \fontsize{8pt}{7.5pt}\texttt{s.eo@yonsei.ac.kr}} &
    \parbox[c]{2.6in}{\centering Hyeonseok Moon$^{4,\dagger}$\\[1pt] \fontsize{8pt}{7.5pt}\texttt{hyns.moon@sookmyung.ac.kr}}
  \end{tabular}%
  }\\[10pt]
  \small
  \makebox[5.28in][c]{%
    \begin{tabular}{c}
      \rule{0pt}{16pt}$^{1}$Korea University \hspace{0.1in} $^{2}$Soongsil University\\[1pt]
      $^{3}$Yonsei University Mirae Campus \hspace{0.1in} $^{4}$Sookmyung Women's University
    \end{tabular}%
  }%
}

\begin{document}

\maketitle
\begingroup
\renewcommand{\thefootnote}{\textdagger}
\footnotetext{Corresponding authors.}
\endgroup
\lhead{Preprint}

\begin{abstract}
    Block drafters for speculative decoding are commonly trained on corpora written by external models, where a single off-policy token invalidates supervision for all subsequent slots in a block. Existing approaches discard these divergent slots, resulting in severe supervision loss. To resolve this problem while preserving the training corpus, we propose a rollout-based training framework that recovers full supervision through two complementary components. The first component, Anchor-Label Relabelling (ALR), replaces corpus labels with distributions from greedy target rollouts, restoring valid supervision across all predicted slots. The second component, In-Rollout Anchors (IRA), places draft blocks directly inside these rollouts to expose the drafter to target-generated context, reusing precomputed rollout features at no additional target cost. Across fixed vision-language and text corpora, our framework increases greedy accepted length by up to 36.5\% over DFlash and consistently outperforms erasing baselines. Notably, a single epoch of our method surpasses the best erase schedules. After three epochs, it matches the acceptance length of training on target-regenerated responses. These results show that our framework provides an effective and compute-efficient approach for training speculative drafters on fixed corpora without modifying the original text. Code is available at \url{https://github.com/js-lee-AI/ALR-IRA}.    \end{abstract}

\section{Introduction}
\label{introduction:main}

In autoregressive language models, speculative decoding reduces generation latency while preserving the target distribution in exact arithmetic \citep{leviathan2022fast,chen2023accelerating}. A small drafter proposes several candidate tokens, and the primary model to be accelerated (the target) verifies them in a single forward pass. The resulting speedup depends on the accepted length, defined as the number of tokens committed in each verification round, and on the relative cost of drafting and verification. Recent work increases accepted length by improving agreement between the drafter and the target, and by training with verification \citep{cai2024medusa,li2024eagle,li2025eagle3,zhou2023distillspec,an2026pard2}. The same framework now also serves vision-language models \citep{kang2025vispec,tong2026sage}, whose instruction-tuning corpora typically consist of responses written by a stronger proprietary model \citep{liu2023visual,chen2024allava,chen2023sharegpt4v}. Such a corpus is off-policy for the target, and its responses often differ from the continuations the target itself would produce.

Block drafters are especially sensitive to this mismatch \citep{an2025pard,chen2026dflash}. Given an anchor token, a block drafter predicts all slots of a block in parallel, and each slot is trained on the corpus token at the same position. The label of a later slot therefore depends on the intermediate corpus tokens, which the drafter does not observe. If one of these tokens differs from the greedy choice of the target, every subsequent label in the block continues a prefix that the target would not generate at inference.

PARD-2 \citep{an2026pard2} addresses this problem by down-weighting such slots. Each slot is weighted by the probability that the target assigns to the preceding corpus tokens in the block, and this weight decays toward zero once the prefix departs from the target distribution. Supervision is thus retained where the corpus agrees with the target and erased where it diverges. Regenerating the corpus with the target avoids this loss of supervision \citep{kim2016sequence,cai2024medusa,li2025eagle3,an2025pard,chen2026dflash}. However, regeneration requires a separate generation stage before training and replaces the original responses, which may need to be kept, for example when only an audited or licensed release can be used for training. In imitation learning, this mismatch is framed as covariate shift, and the standard remedy is to train on newly collected trajectories \citep{ross2010reduction,agarwal2023policy,gu2023minillm,liu2023online}. This approach likewise replaces the underlying training text.

In this paper, we show that a block drafter learns more from greedy continuations of the target than from erasing the off-policy slots of a fixed corpus. The first component of our method, Anchor-Label Relabelling (ALR), leaves the corpus unchanged and updates only the training labels. From each anchor, the target continues the corpus prefix greedily, and each slot is labeled with the target distribution conditioned on the prefix and preceding rollout tokens, erasing none of them. As in standard training, the drafter receives the anchor token and the target features of the corpus prefix. The rollout is computed within the training step by reusing the cached forward pass of the target, avoiding the need to generate or store full responses in advance. For ALR in isolation, a short rollout depth is sufficient.

ALR alone still leaves a residual gap to a drafter trained on responses regenerated by the target at matched lengths. A position-wise analysis shows no measurable gap over the first 16 generated tokens, with the deficit emerging only in later positions. From that point on, the recent context of the drafter at inference consists of text generated by the target, whereas a block anchored strictly in the corpus never observes such context during training. The second component of our method, In-Rollout Anchors (IRA), supplies this target-generated context. IRA keeps the total number of blocks fixed and relocates half of them into the rollouts of the other half. Each relocated block is anchored on a token generated by the target and conditioned on features already computed during the rollout, requiring no additional forward pass of the target. Thus, ALR corrects the labels, while IRA reduces the discrepancy between training and inference contexts.

We evaluate both components on two fixed vision-language corpora and one fixed text corpus. Under greedy decoding, ALR + IRA improves accepted length by up to 36.5\% over DFlash trained on the same data and by up to 8.54\% over the erasing baseline. It also achieves a higher accepted length than the erasing baseline on every evaluated domain and benchmark. These improvements hold for a smaller vision-language target and for a text-only target on seven benchmarks, under both greedy and sampled decoding. With the other training settings matched, a single epoch of ALR + IRA outperforms the best erase schedule we evaluated on either vision-language corpus, at less than twice the cost of one erase epoch. After three epochs of training, ALR + IRA achieves greedy decoding speedups of 2.82$\times$ on ALLaVA and 2.77$\times$ on ShareGPT4V over autoregressive decoding, compared with 2.62$\times$ for erase on both corpora. On ALLaVA, its accepted length is within 0.4\% of drafters trained on responses regenerated by the target, even though the original corpus is kept fixed. The gain of IRA over ALR is largest at positions in held-out captions where an independent drafter fails to predict the first token. 
Ultimately, our findings show that target-generated context complements label relabelling, demonstrating that training-time target rollouts provide a practical alternative to response regeneration when the original corpus must be preserved.
\section{Related Work on Draft Training}
\label{related_work:section}

PARD-2 \citep{an2026pard2} aligns parallel draft training with consecutive token acceptance through confidence-adaptive loss weights. Each weight is the product of the target's probabilities for the preceding corpus tokens within the block. Our erase comparator applies this weighting to knowledge distillation on the fixed corpus. Its labels remain target distributions conditioned on corpus continuations, while unlikely corpus tokens reduce supervision for later slots. ALR obtains these distributions along a greedy target rollout from the same corpus prefix, and IRA also supplies rollout features as drafter context. This extends target alignment to the text that determines both the labels and the drafter's context, while preserving the original corpus.

The EAGLE family generates draft tokens autoregressively using target features \citep{li2024eagle,li2024eagle2}. EAGLE-3 and HASS train on hidden states generated by the drafter to reduce the mismatch between training and multi-step drafting \citep{li2025eagle3,zhang2024learning}. Parallel drafters predict several future positions in one forward pass \citep{stern2018blockwise,cai2024medusa,xiao2024parallelspec,an2025pard}, while semi-autoregressive methods combine sequential and parallel drafting \citep{gao2025falcon}. We use DFlash \citep{chen2026dflash}, a small block-diffusion drafter conditioned on target features. Our method keeps this architecture unchanged while obtaining supervision and context from the target's own continuations.

Knowledge distillation aligns a drafter's output distributions with those of the target \citep{hinton2015distilling,zhou2023distillspec}. On-policy distillation uses student-generated continuations \citep{agarwal2023policy,gu2023minillm}, while online speculative decoding adapts the drafter using feedback collected during serving \citep{liu2023online}. Model-generated responses also support sequence-level and reasoning distillation \citep{kim2016sequence,lee2026answerconditioned}, and several draft-training methods use target-generated responses \citep{cai2024medusa,goel2024direct,zafrir2025fastdraft,hong2025training}. Regenerating the corpus provides target-written supervision and context, but replaces the original responses. When these responses must be preserved, computing target distributions along them still leaves supervision conditioned on continuations that may differ from the target's own. Our ALR + IRA addresses this mismatch with short target rollouts from corpus prefixes during training, supplying both supervision and context without rewriting the corpus.

For vision-language targets, prior work has developed both visual processing and training strategies. ViSpec \citep{kang2025vispec} compresses image tokens with a vision adaptor, DREAM \citep{hu2025dream} fuses target features through cross-attention, and MSD \citep{lin2025speculative} processes visual and text tokens differently and gradually introduces multimodal data after text-only training. ViSpec, DREAM, and MASSV \citep{ganesan2025massv} also train on target-generated responses. Our ALR + IRA changes training labels and anchor context through short target rollouts on a fixed corpus, without introducing image-specific components.

\section{Relabelling Anchor Blocks with the Target's Own Rollout}
\label{method:sec}

\begin{figure}[t]
\centering
\includegraphics[width=\linewidth]{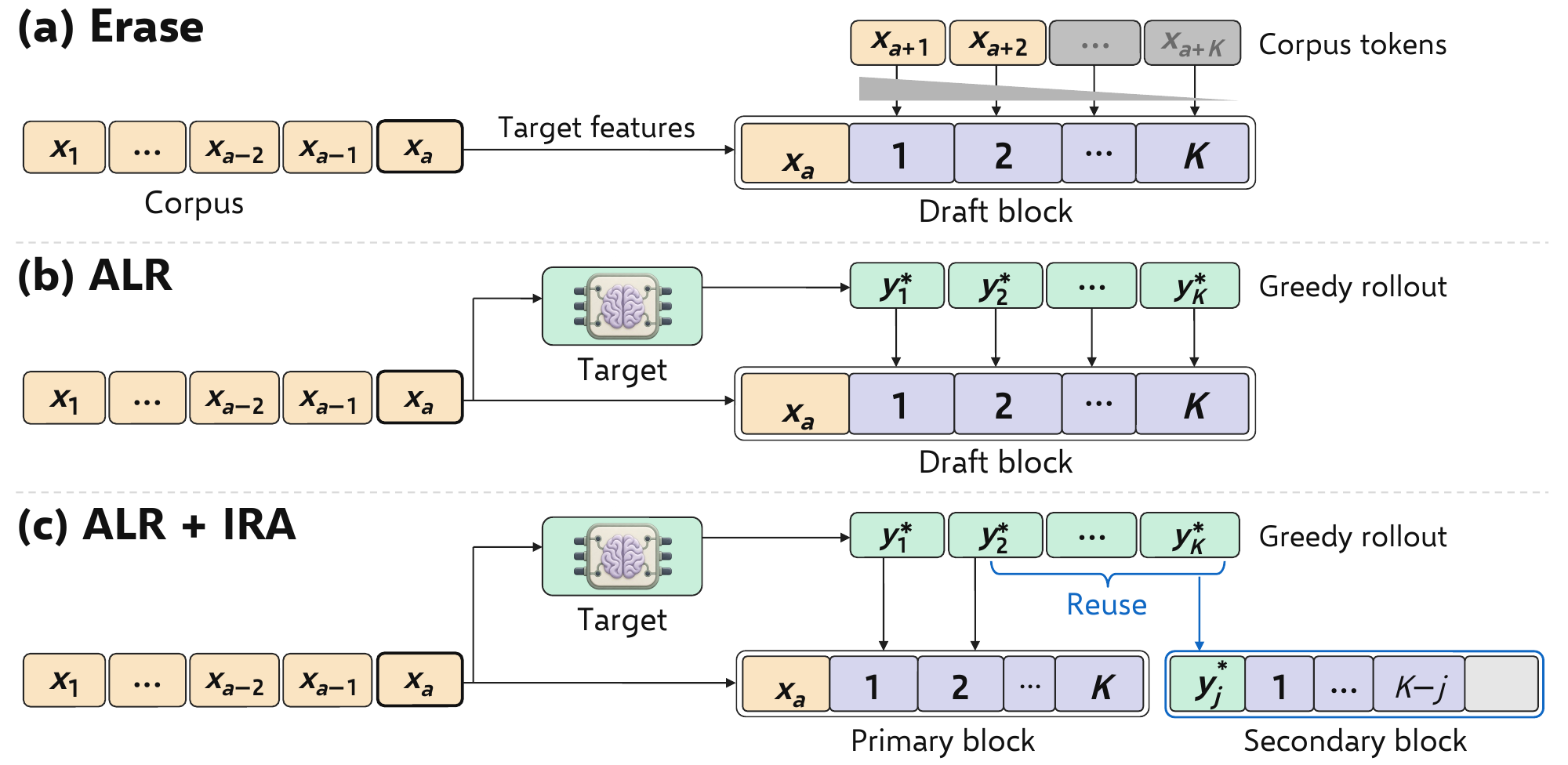}
\caption{Label construction from a fixed corpus. Amber and green denote corpus and target-rollout tokens, respectively. Purple cells are draft slots, and black downward arrows supply the target distributions predicting the tokens above them. Erase down-weights corpus-conditioned labels, whereas ALR uses rollout-conditioned labels. IRA retains the corpus context and reuses rollout features and labels in a secondary block without another target pass. Its gray tail is unsupervised.}
\label{method:labels}
\end{figure}

For a response sequence $x$ from the training corpus and an index $a$, an \emph{anchor block} spans $K+1$ slots beginning at position $a$. Slot 0 contains the anchor token $x_a$, and the drafter predicts slots 1 through $K$ in parallel conditioned on the target hidden representations up to position $a$. Under the standard corpus objective, slot $k$ receives the one-hot target $x_{a+k}$. Following DFlash \citep{chen2026dflash}, we set $K=15$ with a five-layer architecture that receives target features from layers 1, 9, 17, 25, and 33, sampling 128 anchors per training sequence. All drafters are initialized from the publicly released text-only DFlash head corresponding to the target base model without architectural changes. Appendix~\ref{app:training} details these checkpoints and additional training hyperparameters. The compared objectives vary the target distribution and loss weight assigned to each slot, while IRA additionally relocates half of the draft blocks into rollout trajectories. Figure~\ref{method:labels} illustrates label construction across erase, ALR, and ALR + IRA. We denote the target next-token distribution by $p_T$, omitting prompt and image conditioning tokens from the notation for simplicity.

\subsection{Labels from the corpus prefix}

Every objective beyond the corpus baseline replaces the one-hot target with a distribution $\pi_{a,k}$ and assigns a slot weight $\omega_{a,k}$,
\begin{equation}
\mathcal{L}(\theta)=\frac{\sum_{a\in\mathcal{A}}\sum_{k=1}^{K}\omega_{a,k}\,\ell\big(\pi_{a,k},\,q_\theta^{k}(\cdot\mid a)\big)}{\sum_{a\in\mathcal{A}}\sum_{k=1}^{K}\omega_{a,k}},
\qquad w_k=e^{-(k-1)/\gamma},
\label{method:loss}
\end{equation}
where $\mathcal{A}$ denotes the set of anchors in a batch, $q_\theta^{k}(\cdot\mid a)$ is the drafter prediction for slot $k$ at anchor $a$, and $\ell$ evaluates soft cross-entropy against the target distribution restricted to the top 8 tokens with the residual probability mass grouped into a tail bin. Normalization by the total weight allows gating $\omega_{a,k}$ to redistribute supervision across slots without shrinking the overall loss scale. All distillation objectives incorporate the decay envelope $w_k$ into $\omega_{a,k}$, setting $\gamma=2$ by default. For the fixed-corpus baseline, we apply knowledge distillation (KD) \citep{zhou2023distillspec} using $\pi_{a,k}=p_T(\cdot\mid x_{<a+k})$ and $\omega_{a,k}=w_k$. This formulation leaves labels conditioned on corpus continuations $x_{a+1},\dots,x_{a+k-1}$ that the target may not generate. During inference, a block is drafted over text generated by the target itself, and slot $k$ contributes to the accepted length only when the target accepts all preceding candidate tokens.

\paragraph{Erasing off-policy slots.}
Under the erase objective, we retain the distillation distribution and scale the slot weight by a survival gate,
\begin{equation}
g_{a,k}=\prod_{1\le j<k}p_T\big(x_{a+j}\mid x_{<a+j}\big),\qquad \omega_{a,k}=w_k\,g_{a,k},
\label{method:gate}
\end{equation}
adapting the confidence-adaptive survival weighting of PARD-2 \citep{an2026pard2} to distillation. The gate equals 1 at slot 1, decreases monotonically along the block, and down-weights slots whose preceding corpus tokens have low probability under the target. A hard variant replaces each factor with the indicator $\mathbf{1}[x_{a+j}=\hat y_{a+j}]$, where $\hat y_{a+j}$ denotes the target greedy prediction after $x_{<a+j}$, discarding all subsequent slots upon the first mismatch. We adopt the soft survival formulation as the primary erase comparator throughout this work. In both variants, supervision remains conditioned on the corpus continuation, with only the loss weights changed.

\subsection{Labels from the target's rollout}

Anchor-label relabelling (ALR) retains the corpus context while substituting the corpus continuation with greedy rollouts generated by the target. Starting from each anchor, the target extends $x_{\le a}$ greedily, providing slot $k$ with the target distribution conditioned on the preceding $k-1$ rollout tokens,
\begin{equation}
y^*_i=\arg\max_v\, p_T\big(v\mid x_{\le a},y^*_{<i}\big),
\qquad
\pi_{a,k}=p_T\big(\cdot\mid x_{\le a},y^*_{<k}\big),
\label{method:alr}
\end{equation}
where the rollout depth $R$ counts the greedy tokens fed back to the target, yielding up to $R+1$ slot distributions. We set $R=K-1$ to provide supervision across all $K$ slots whenever the target does not terminate early. In-rollout anchors introduced below require this full depth and are never evaluated with shorter rollouts. Consequently, experimental configurations with $R<K-1$ evaluate ALR in isolation, leaving slots beyond $R+1$ unconstrained. Slots following early sequence termination are masked, while remaining slots retain $\omega_{a,k}=w_k$, ensuring that full-depth ALR preserves supervision across every active position. 

On a greedy target rollout, the hard variant of the gate in Eq.~\eqref{method:gate} equals 1 across all active slots. The soft gate can still decrease because greedy tokens need not have probability one, so ALR omits it. Slot 1 receives identical supervision under KD, erase, and ALR, as $y^*_{<1}$ is empty. For subsequent slots, each label aligns with the greedy target trajectory verified during greedy decoding. Since slot $k$ contributes to the accepted length only when slots $1$ through $k-1$ match $y^*_1,\dots,y^*_{k-1}$, the ALR distribution matches the verification distribution whenever slot $k$ contributes to the accepted length. Drafter inputs, consisting of the anchor token and target representations up to $a$, remain unchanged from the standard corpus objective.

Table~\ref{analysis:regeneration}b compares ALR against a drafter trained with the same KD loss on target-regenerated responses truncated to the corpus lengths. On COCO captions, ALR matches this regeneration reference over the initial 16 generated tokens before exhibiting a performance deficit that subsequently plateaus. Every anchor in a regenerated response follows context generated by the target in preceding tokens. In contrast, an ALR block encounters target-generated tokens only in its supervision labels, as drafter features condition strictly on the corpus prefix up to $a$. We attribute this gap to the absence of target-generated context, motivating a secondary anchor block described next.

\paragraph{In-rollout anchors.}
In-rollout anchors (IRA) preserve ALR's anchor draw and total block count $m=\min(v,128)$, where $v$ is the number of valid anchor positions in the sample, while partitioning the blocks into two subsets. The $\lceil m/2\rceil$ primary blocks follow the standard ALR formulation in Eq.~\eqref{method:alr}. Each of the $\lfloor m/2\rfloor$ secondary blocks is positioned within a primary rollout at an offset $j\sim\mathcal{U}\{2,\dots,K-1\}$. Its conditioning context concatenates the target representations of the corpus prefix up to $a$ with the intermediate representations stored for $y^*_1,\dots,y^*_{j-1}$. Slot 0 is assigned the target-generated token $y^*_j$, and slot $k$ inherits the primary block label at position $j+k$,
\begin{equation}
\pi^{(j)}_{a,k}=\pi_{a,j+k}=p_T\big(\cdot\mid x_{\le a},y^*_{<j+k}\big),\qquad \omega^{(j)}_{a,k}=w_k\,\mathbf{1}[\,j+k\le K\,].
\label{method:ira}
\end{equation}
A secondary block provides an anchor whose immediate prefix consists of target-generated text, supervising up to $K-j$ slots, averaging seven slots prior to sequence termination. This design incurs no additional forward passes over the target model, as representations for $y^*_{<j}$ and target distributions past index $j$ are already computed during the primary rollout.
Only the $\lceil m/2\rceil$ primary blocks initiate rollout generation, compared to all $m$ blocks in ALR. By reusing rollout features and distributions from primary blocks, IRA introduces secondary anchors without increasing target forward computation.

\section{Accepted Length and Speedup on a Fixed Corpus}
\label{results:sec}

\paragraph{Experimental setup.} For vision-language experiments, we adopt Qwen3-VL-8B-Instruct \citep{bai2025qwen3vl} as the target model and pair it with a DFlash block drafter \citep{chen2026dflash} initialized from the public text-only DFlash head. We train drafters on either the \texttt{allava\_laion} split of ALLaVA \citep{chen2024allava} or ShareGPT4V captions \citep{chen2023sharegpt4v}, matching both corpora to an identical count of image-text pairs. For the text-only target Qwen3-4B \citep{yang2025qwen3}, we train all four methods on the same UltraChat subset \citep{ding2023ultrachat}, keeping all three training corpora strictly fixed. We evaluate vision-language drafting on COCO captioning \citep{lin2014coco}, TextVQA \citep{singh2019textvqa}, and DocVQA \citep{mathew2021docvqa}, and assess text-only drafting across seven benchmarks spanning instruction following, mathematical reasoning, and code generation.

\paragraph{Evaluation metrics.} We report mean acceptance length $\tau$ (MAT), counting accepted draft tokens plus one target token per verification step, alongside wall-clock speedup over autoregressive decoding (AR). To compute aggregate acceptance lengths, we take the geometric mean across evaluation tasks within each training seed and then average across seeds. For vision-language benchmarks, we evaluate the first 100 prompts per domain, capping generation at 256 new tokens across all tasks. We examine both greedy decoding ($T{=}0$) and stochastic sampling ($T{=}1$). We provide detailed evaluation protocols in Appendices~\ref{app:eval} and~\ref{app:text}. Given that bf16 tree verification trajectories can vary slightly across GPU architectures, we measure acceptance length for the primary vision-language target on RTX A6000 hardware and evaluate the smaller vision-language and text-only targets on H100 GPUs. We measure the speedups in Table~\ref{results:tab-mat} on H100 GPUs. In Appendix~\ref{app:seeds}, we document training replicates and sensitivity analyses, and in Appendix~\ref{app:determinism}, we define the formal comparison rule for vision-language evaluations.

\subsection{Accepted length across targets}
\label{results:mat}

\begin{table}[t]
\centering
\caption{Speedup over autoregressive decoding (AR) and mean acceptance length $\tau$ after three epochs on each fixed training corpus. Vision values summarize COCO captioning, TextVQA and DocVQA. Text rows identify evaluation benchmarks. Aggregates are geometric means across benchmarks, averaged over available replicates.}
\label{results:tab-mat}
\fontsize{8.5}{9.5}\selectfont
\setlength{\tabcolsep}{2pt}
\setlength{\aboverulesep}{0.25ex}
\setlength{\belowrulesep}{0.4ex}
\setlength{\parskip}{0pt}
\begin{tabularx}{\linewidth}{l*{6}{>{\centering\arraybackslash}X}}
\toprule
\grouprow{7}{Vision} \\
Target & \multicolumn{4}{c}{Qwen3-VL-8B} & \multicolumn{2}{c}{Qwen3-VL-4B} \\
\cmidrule(lr){2-5}\cmidrule(lr){6-7}
Train corpus & \multicolumn{2}{c}{ALLaVA} & \multicolumn{2}{c}{ShareGPT4V} & \multicolumn{2}{c}{ALLaVA} \\
\cmidrule(lr){2-3}\cmidrule(lr){4-5}\cmidrule(lr){6-7}
Method & Speedup & $\tau$ & Speedup & $\tau$ & Speedup & $\tau$ \\
\midrule
\grouprow{7}{$T{=}0$} \\
DFlash & $2.09\times$ & $2.88$ & $2.01\times$ & $2.80$ & $2.19\times$ & $2.92$ \\
Erase & $2.62\times$ & $3.54$ & $2.62\times$ & $3.57$ & $2.62\times$ & $3.56$ \\
ALR & $2.68\times$ & $3.70$ & $\mathbf{2.80}\times$ & $3.67$ & $2.65\times$ & $3.63$ \\
\rowcolor{ourtint}
ALR + IRA & $\mathbf{2.82}\times$ & $\mathbf{3.84}$ & $2.77\times$ & $\mathbf{3.83}$ & $\mathbf{2.77}\times$ & $\mathbf{3.81}$ \\
\midrule
\grouprow{7}{$T{=}1$} \\
DFlash & $2.04\times$ & $2.79$ & $1.90\times$ & $2.71$ & $2.04\times$ & $2.78$ \\
Erase & $2.49\times$ & $3.39$ & $2.54\times$ & $3.39$ & $2.45\times$ & $3.33$ \\
ALR & $2.55\times$ & $3.50$ & $2.54\times$ & $3.48$ & $2.53\times$ & $3.41$ \\
\rowcolor{ourtint}
ALR + IRA & $\mathbf{2.68}\times$ & $\mathbf{3.65}$ & $\mathbf{2.58}\times$ & $\mathbf{3.60}$ & $\mathbf{2.67}\times$ & $\mathbf{3.53}$ \\
\bottomrule
\end{tabularx}
\par
\begin{tabularx}{\linewidth}{l*{6}{>{\centering\arraybackslash}X}*{2}{>{\columncolor{ourtint}\centering\arraybackslash}X}}
\toprule
\grouprow{9}{Text benchmarks: Qwen3-4B} \\
 & \multicolumn{2}{c}{DFlash} & \multicolumn{2}{c}{Erase} & \multicolumn{2}{c}{ALR} & \multicolumn{2}{c}{\cellcolor{ourtint}ALR + IRA} \\
\cmidrule(lr){2-3}\cmidrule(lr){4-5}\cmidrule(lr){6-7}\cmidrule(lr){8-9}
Benchmark & Speedup & $\tau$ & Speedup & $\tau$ & Speedup & $\tau$ & Speedup & $\tau$ \\
\midrule
\grouprow{9}{$T{=}0$} \\
MT-Bench & $1.82\times$ & $2.50$ & $2.07\times$ & $2.85$ & $2.09\times$ & $2.94$ & $\mathbf{2.13}\times$ & $\mathbf{2.99}$ \\
Alpaca & $1.56\times$ & $2.17$ & $1.88\times$ & $2.52$ & $1.87\times$ & $2.58$ & $\mathbf{1.92}\times$ & $\mathbf{2.62}$ \\
GSM8K & $2.79\times$ & $3.96$ & $3.16\times$ & $4.48$ & $3.38\times$ & $4.76$ & $\mathbf{3.55}\times$ & $\mathbf{4.83}$ \\
AIME24 & $2.61\times$ & $3.69$ & $2.85\times$ & $4.06$ & $2.91\times$ & $4.20$ & $\mathbf{2.97}\times$ & $\mathbf{4.27}$ \\
AIME25 & $2.76\times$ & $4.07$ & $3.01\times$ & $4.34$ & $2.94\times$ & $4.53$ & $\mathbf{3.14}\times$ & $\mathbf{4.66}$ \\
HumanEval & $2.98\times$ & $4.22$ & $3.46\times$ & $4.92$ & $3.56\times$ & $4.98$ & $\mathbf{3.63}\times$ & $\mathbf{5.02}$ \\
LiveCodeBench & $2.33\times$ & $3.25$ & $2.64\times$ & $3.72$ & $2.62\times$ & $3.81$ & $\mathbf{2.73}\times$ & $\mathbf{3.84}$ \\
\midrule
Avg. (7 tasks) & $2.35\times$ & $3.32$ & $2.67\times$ & $3.75$ & $2.70\times$ & $3.87$ & $\mathbf{2.80}\times$ & $\mathbf{3.93}$ \\
\midrule
\grouprow{9}{$T{=}1$} \\
MT-Bench & $1.79\times$ & $2.42$ & $1.97\times$ & $2.72$ & $2.06\times$ & $2.79$ & $\mathbf{2.15}\times$ & $\mathbf{2.83}$ \\
Alpaca & $1.52\times$ & $2.16$ & $1.72\times$ & $2.47$ & $1.78\times$ & $2.54$ & $\mathbf{1.80}\times$ & $\mathbf{2.57}$ \\
GSM8K & $2.71\times$ & $3.78$ & $3.01\times$ & $4.26$ & $3.19\times$ & $4.50$ & $\mathbf{3.29}\times$ & $\mathbf{4.55}$ \\
AIME24 & $2.41\times$ & $3.35$ & $2.49\times$ & $3.54$ & $2.58\times$ & $\mathbf{3.66}$ & $\mathbf{2.59}\times$ & $3.65$ \\
AIME25 & $2.40\times$ & $3.49$ & $2.62\times$ & $3.77$ & $2.46\times$ & $3.85$ & $\mathbf{2.66}\times$ & $\mathbf{3.96}$ \\
HumanEval & $2.89\times$ & $3.97$ & $\mathbf{3.22}\times$ & $4.57$ & $3.19\times$ & $4.63$ & $3.20\times$ & $\mathbf{4.66}$ \\
LiveCodeBench & $2.20\times$ & $3.00$ & $2.44\times$ & $3.38$ & $2.42\times$ & $3.47$ & $\mathbf{2.52}\times$ & $\mathbf{3.49}$ \\
\midrule
Avg. (7 tasks) & $2.23\times$ & $3.10$ & $2.45\times$ & $3.46$ & $2.48\times$ & $3.56$ & $\mathbf{2.55}\times$ & $\mathbf{3.60}$ \\
\bottomrule
\end{tabularx}
\end{table}

\begin{figure}[t]
\centering
\setlength{\abovecaptionskip}{0pt}
\includegraphics{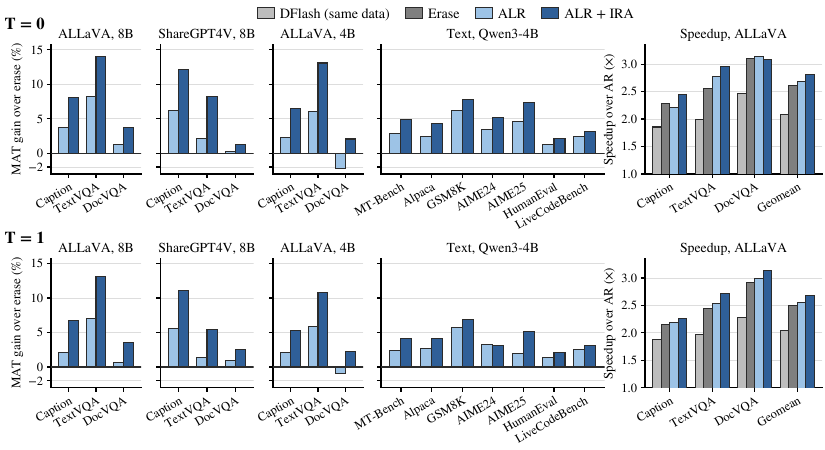}
\caption{MAT gain over erase across vision-language and text benchmarks, and ALLaVA decoding speedup over autoregressive decoding after three training epochs. The top and bottom rows show $T{=}0$ and $T{=}1$, respectively. Gains use means across training replicates.}
\label{results:fig}
\end{figure}

Table~\ref{results:tab-mat} summarizes accepted lengths and decoding speedups following three epochs of training on each fixed corpus. Under greedy decoding ($T{=}0$), ALR + IRA attains the highest accepted length across all three vision-language configurations. On both 8B corpora, ALR consistently improves upon the soft survival-weighted erase baseline from PARD-2 \citep{an2026pard2}, with IRA providing an additional boost that reaches an 8.54\% margin over erase on ALLaVA. Both steps clear the registered comparison threshold on these datasets. As reported in Appendix~\ref{app:robust}, soft gating outperforms matched ungated distillation, establishing erase as a competitive baseline. The relative ordering across methods remains consistent within each domain of both 8B corpora (Table~\ref{app:vision-domains}), where DocVQA exhibits the narrowest gap. Under stochastic sampling ($T{=}1$), the aggregate ranking is preserved even though the training labels derive from greedy target rollouts. Figure~\ref{results:fig} illustrates these benchmark-level gains over erase under both temperature settings. Furthermore, Table~\ref{app:tab-robust} documents a 5.90\% aggregate accepted-length improvement over erase for ALR + IRA under chain verification at $T{=}0$ on ALLaVA, demonstrating that the gain does not rely on tree branching.

\paragraph{ALR + IRA outperforms DFlash and the best erase schedules.}
ALR + IRA improves accepted length over DFlash trained on identical data by up to 36.5\% at $T{=}0$ and 33.1\% at $T{=}1$ across both 8B settings (Table~\ref{results:tab-mat}). Across training horizons, the combined framework also surpasses the strongest evaluated erase schedule on each corpus at $T{=}0$ (Table~\ref{app:epoch-schedules}). On ShareGPT4V, where the top baseline reference corresponds to single-epoch erase, ALR + IRA maintains a 5.09\% lead.

\paragraph{The gain repeats with a second target.}
These improvements generalize to an alternative target model, where ALR + IRA exceeds erase by 7.13\% in accepted length at $T{=}0$ using Qwen3-VL-4B-Instruct (Table~\ref{results:tab-mat}). On DocVQA, ALR in isolation falls below erase at both decoding temperatures, whereas ALR + IRA rises above it (Figure~\ref{results:fig}). This reversal confirms the distinct value of IRA in regimes where relabelling alone does not overcome erase.

\subsection{Decoding speedup}
\label{results:speed}

Figure~\ref{results:fig} shows that ALR + IRA decodes faster than erase on ALLaVA at both temperatures: 2.82$\times$ versus 2.62$\times$ over AR at $T{=}0$, and 2.68$\times$ versus 2.49$\times$ at $T{=}1$. At $T{=}0$, captioning and TextVQA drive the speed gain while DocVQA is nearly tied despite higher accepted length. At $T{=}1$, ALR + IRA is faster than erase in all three domains. Table~\ref{results:tab-mat} shows the speed advantage also holds on ShareGPT4V under sampling, and Table~\ref{app:tab-speed} reports a 4.1\% advantage over its one-epoch erase reference at $T{=}0$. Appendix~\ref{app:eagle-h100} reports the separate EAGLE-3 comparison at both temperatures.

\paragraph{The gains extend to a text-only target.}
Table~\ref{results:tab-mat} shows that ALR + IRA has the highest aggregate accepted length and decoding speedup across the seven text benchmarks at both temperatures. Its accepted-length gain over erase reaches 4.96\% at $T{=}0$ and 4.10\% at $T{=}1$. Table~\ref{app:text-timing} confirms aggregate speed gains over erase of 4.62\% at $T{=}0$ and 3.89\% at $T{=}1$ under repeated A100 timing on a fixed text subset.

\raggedbottom
\section{What the Recovered Supervision Costs and Where It Acts}
\label{analysis:sec}

\noindent
\begin{minipage}{\textwidth}
\setlength{\intextsep}{2pt}
\setlength{\columnsep}{7pt}
\begin{wraptable}{R}{0.45\textwidth}
\centering
\setlength{\abovecaptionskip}{0pt}
\setlength{\belowcaptionskip}{0pt}
\setlength{\abovetopsep}{2pt}
\caption{One-epoch training and the best erase schedules C at $T{=}0$. ALR uses $R{=}4$, and ALR + IRA uses $R{=}14$. Mean $\pm$ SD.}
\label{analysis:schedules}
\fontsize{7}{8}\selectfont
\setlength{\tabcolsep}{2.3pt}
\setlength{\aboverulesep}{0.2ex}
\setlength{\belowrulesep}{0.3ex}
\begin{tabularx}{\linewidth}{l*{2}{>{\centering\arraybackslash}X}c}
\toprule
Method & Epochs & Time/C & $\tau$ \\
\midrule
\grouprow{4}{ALLaVA} \\
Erase & 1 & 0.33 & $3.533_{\pm0.006}$ \\
ALR & 1 & 0.42 & $3.649_{\pm0.014}$ \\
\rowcolor{ourtint}
ALR + IRA & 1 & 0.59 & $\mathbf{3.728}_{\pm0.005}$ \\
Erase (C) & 3 & 1.00 & $3.539_{\pm0.010}$ \\
\midrule
\grouprow{4}{ShareGPT4V} \\
Erase (C) & 1 & 1.00 & $3.640_{\pm0.002}$ \\
ALR & 1 & 1.28 & $3.695_{\pm0.011}$ \\
\rowcolor{ourtint}
ALR + IRA & 1 & 1.77 & $\mathbf{3.778}_{\pm0.003}$ \\
\bottomrule
\end{tabularx}

\end{wraptable}

Table~\ref{analysis:schedules} pairs training time with accepted length for one-epoch ALR and ALR + IRA and for C, the best erase schedule measured on each vision-language corpus. This comparison accounts for target-rollout compute alongside data exposure. Appendix~\ref{app:schedules} gives the full epoch, envelope and rollout-depth sweeps.

\paragraph{One epoch exceeds the best erase schedules.}
\mbox{Table~\ref{analysis:schedules}} shows that a single epoch of ALR + IRA exceeds the best erase schedule on both corpora, while costing less than two epochs of erase. On ALLaVA, it achieves this higher accepted length in 41\% less training time than the best erase schedule.
\par\WFclear
\end{minipage}
\par

\noindent
\begin{minipage}{\textwidth}
\setlength{\intextsep}{0pt}
\setlength{\columnsep}{8pt}
\begin{wrapfigure}{R}{0.54\textwidth}
\centering
\setlength{\abovecaptionskip}{0pt}
\includegraphics{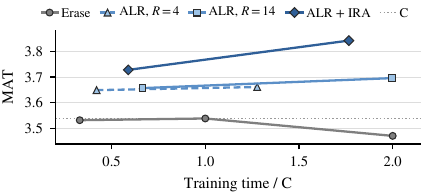}
\caption{ALLaVA training cost at $T{=}0$. Points show one and three epochs, plus six for erase. Time is relative to C, whose MAT the dotted line marks.}
\label{analysis:fig-cost}
\end{wrapfigure}
\paragraph{Additional training preserves the advantage.}
Figure~\ref{analysis:fig-cost} shows that ALR + IRA improves from one to three epochs, while training erase for six epochs does not close the gap. At both recorded epoch counts, ALR + IRA achieves higher accepted length than full-depth ALR in less training time. Its secondary blocks reuse features and labels from the primary rollouts. This reuse reduces rollout computation while providing target-generated context. The trajectories suggest that recovering supervision is more effective than additional passes over the fixed corpus.
\par\WFclear
\end{minipage}
\par

\noindent
\begin{minipage}{\textwidth}
\setlength{\intextsep}{0pt}
\setlength{\columnsep}{8pt}
\begin{wraptable}{R}{0.37\textwidth}
\centering
\setlength{\abovecaptionskip}{0pt}
\setlength{\belowcaptionskip}{0pt}
\setlength{\abovetopsep}{1pt}
\fontsize{9}{10.5}\selectfont
\captionof{table}{ALR depth after three epochs at $T{=}0$. Mean $\pm$ SD.}
\label{analysis:rollout-depth-summary}
\fontsize{7}{8}\selectfont
\setlength{\tabcolsep}{3pt}
\setlength{\aboverulesep}{0.2ex}
\setlength{\belowrulesep}{0.3ex}
\begin{tabularx}{\linewidth}{>{\centering\arraybackslash}X>{\centering\arraybackslash}X}
\toprule
Rollout depth $R$ & MAT $\uparrow$ \\
\midrule
\grouprow{2}{ALLaVA} \\
14 & $3.696_{\pm0.005}$ \\
6 & $3.682_{\pm0.004}$ \\
4 & $3.661_{\pm0.023}$ \\
2 & $3.597_{\pm0.017}$ \\
\midrule
\grouprow{2}{ShareGPT4V} \\
14 & $3.673_{\pm0.007}$ \\
4 & $3.659_{\pm0.014}$ \\
\bottomrule
\end{tabularx}

\end{wraptable}
\paragraph{Short rollouts suffice for ALR.}
\mbox{Table~\ref{analysis:rollout-depth-summary}} shows that $R=4$ is the shortest tested ALR depth that is non-inferior to $R=14$ after three epochs on ALLaVA. It also meets the comparison criterion on ShareGPT4V and trains faster than $R=14$ on both corpora, supporting its use for the low-cost ALR setting in Table~\ref{analysis:schedules}. The criterion is defined in Appendix~\ref{app:determinism}, and \mbox{Table~\ref{app:rollout-depths}} gives the full sweep. The main three-epoch comparison keeps $R=14$ for both ALR and ALR + IRA. IRA needs that depth because it draws secondary anchors at offsets up to $K-1$ inside the rollout.
\par\WFclear
\end{minipage}
\par

\begingroup
\setlength{\intextsep}{3pt}
\setlength{\columnsep}{8pt}
\begin{wraptable}{r}{0.485\textwidth}
\centering
\setlength{\abovecaptionskip}{0pt}
\setlength{\belowcaptionskip}{1pt}
\setlength{\abovetopsep}{2pt}
\caption{Length-matched target regeneration on the same ALLaVA prompts after three epochs at $T{=}0$. (a) Overall MAT with Qwen3-VL-8B. (b) MAT gains over ALR by round-start token offset on COCO captions.}
\label{analysis:regeneration}
\fontsize{8}{9.5}\selectfont
\setlength{\tabcolsep}{1.3pt}
\begin{tabular}{lrrrrr}
\toprule
& (a) Overall & \multicolumn{4}{c}{(b) COCO gain over ALR (\%)} \\
\cmidrule(lr){2-2}\cmidrule(lr){3-6}
Method & MAT $\uparrow$ & 0--15 & 16--63 & 64--127 & 128+ \\
\midrule
ALR & 3.696 & 0.00 & 0.00 & 0.00 & 0.00 \\
\rowcolor{ourtint}
ALR + IRA & 3.841 & $+1.32$ & $+5.60$ & $+5.15$ & $+3.57$ \\
\grouprow{6}{Target regeneration} \\
\quad Truncated & 3.833 & $+1.49$ & $+5.61$ & $+4.51$ & $+4.72$ \\
\quad Loss window & \textbf{3.856} & $+3.60$ & $+7.21$ & $+5.12$ & $+5.65$ \\
\bottomrule
\end{tabular}
\end{wraptable}

\paragraph{ALR + IRA matches length-matched regeneration on a fixed corpus.}
Table~\ref{analysis:regeneration}-(a) shows that, after three epochs on fixed ALLaVA, our ALR + IRA comes within 0.4\% of both target-regeneration references in aggregate accepted length. Response regeneration aligns the drafter's training text with the target \citep{cai2024medusa,zhou2023distillspec}; our two references use the same distillation loss and control supervised response length by truncation or a loss window. This agreement suggests that rollout labels and in-rollout context can recover the accepted-length benefit of length-controlled regeneration without replacing the training corpus.
\par\WFclear
\endgroup

\suppressfloats[t]
\begin{figure}[t]
\centering
\setlength{\abovecaptionskip}{0pt}
\includegraphics{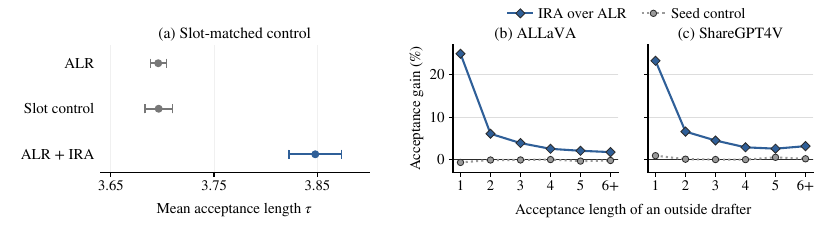}
\caption{IRA controls. (a) Slot-matched control on ALLaVA at $T{=}0$. Error bars show sample SD across matched training replicates. (b,c) IRA's acceptance-length gain over ALR on held-out captions, grouped by an independent drafter's acceptance length, with an ALR seed control.}
\label{analysis:fig-ira}
\end{figure}

\paragraph{IRA gains beyond changing slot weights.}
Figure~\ref{analysis:fig-ira}-(a) shows that ALR + IRA achieves 4.10\% higher accepted length than the slot-matched control on ALLaVA after three epochs. The control matches IRA's block count, offsets and slot weights while keeping all blocks at corpus anchors, yet performs similarly to ALR. This supports placing anchors inside target rollouts to train the drafter on both recent target-generated context and its continuation labels.

\paragraph{IRA's gain concentrates where drafting is hard.}
Figure~\ref{analysis:fig-ira}-(b)--(c) groups drafting offsets in held-out captions, disjoint from the main evaluation, by the acceptance length of an independent drafter trained on the other corpus. This drafter is the compared arms' common parent. On both corpora, IRA's gain is largest where that drafter misses the first token, whereas the seed control stays flat and the gain is smaller where it accepts five or more tokens. Appendix~\ref{app:robust} gives additional controls.

\paragraph{IRA's benefit grows beyond the opening tokens.}
Table~\ref{analysis:regeneration}-(b) shows that IRA's gain over ALR is larger at later positions in COCO captions. The truncated regeneration reference shows a related pattern: its advantage over ALR grows after the first 16 caption tokens and then levels off. Together, these patterns support exposing the drafter to the target-generated context that accumulates during decoding. The positional rise also persists when all drafters are scored on fixed text, while an ALR seed control shows no such rise, as detailed in Appendix~\ref{app:robust}.

\noindent
\begin{minipage}{\textwidth}
\setlength{\intextsep}{0pt}
\setlength{\columnsep}{8pt}
\begin{wraptable}{R}{0.50\textwidth}
\begingroup
\centering
\setlength{\abovecaptionskip}{0pt}
\setlength{\belowcaptionskip}{0pt}
\setlength{\abovetopsep}{2pt}
\caption{Label source and survival weighting after three epochs on fixed UltraChat. MAT aggregates seven text benchmarks. Entries report mean $\pm$ SD across matched training replicates.}
\label{analysis:label-controls}
\fontsize{7}{8.2}\selectfont
\setlength{\tabcolsep}{1.5pt}
\renewcommand{\arraystretch}{1.0}
\begin{tabularx}{\linewidth}{Xccc}
\toprule
Label source & Soft gate & $T{=}0$ & $T{=}1$ \\
\midrule
Corpus (KD) & Off & $3.586\,\pm\,0.0004$ & $3.306\,\pm\,0.005$ \\
Corpus (erase) & On & $3.758\,\pm\,0.015$ & $3.483\,\pm\,0.027$ \\
Target rollout & On & $3.886\,\pm\,0.001$ & $3.583\,\pm\,0.007$ \\
\rowcolor{ourtint}
Target rollout (ALR) & Off & $3.884\,\pm\,0.008$ & $3.580\,\pm\,0.035$ \\
\bottomrule
\end{tabularx}
\par\endgroup

\end{wraptable}

\paragraph{Rollout labels improve acceptance under either weighting rule.}
Table~\ref{analysis:label-controls} shows that rollout-conditioned labels improve aggregate accepted length under both choices of survival weighting on the fixed text corpus. The four arms keep the training recipe and corpus anchors fixed, separating label construction from the use of the gate. Survival weighting helps corpus-conditioned labels but adds little to rollout-conditioned labels at either decoding temperature. The gain therefore persists when the weighting rule is retained, supporting the use of the target's continuation to construct later-slot supervision.
\par\WFclear
\end{minipage}
\par

\flushbottom

\section{Conclusion}
\label{conclusion:main}

In this work, we demonstrated that block drafters trained on fixed off-policy corpora benefit substantially more from supervision along target greedy rollouts than from pruning divergent slots. Our framework recovers this missing supervision through two complementary mechanisms. Anchor-Label Relabelling (ALR) derives valid target distributions along greedy rollouts across all draft positions, while In-Rollout Anchors (IRA) position secondary blocks within these trajectories to expose the drafter to target-generated context. By reusing intermediate representations computed during the primary rollout, IRA supplies this recent context without incurring additional forward passes over the target model. Across two fixed vision-language corpora, the integrated framework improves greedy accepted length by up to 36.5\% over DFlash trained on identical data and consistently outperforms the strongest evaluated erase schedules. These empirical gains translate into concrete decoding speedups, generalize to a smaller vision-language target, and transfer directly to text-only benchmarks. Crucially, on ALLaVA, ALR + IRA matches the acceptance length of training on target-regenerated responses, confirming that rollout-based training provides a practical and compute-efficient alternative when the original training corpus must remain intact. Ultimately, our findings establish target rollouts as a robust foundation for off-policy draft training, and we plan to extend this framework to broader drafter architectures in future work.

\section*{Ethics Statement}

This work trains draft models for speculative decoding using public text and image-text corpora and public language and vision-language models. It involves no human-subject experiments or new data collection. Speculative decoding preserves the target distribution in exact arithmetic, and the method inherits the target's biases and failure modes \citep{lee2022norose}. The vision-language training costs are reported in Section~\ref{analysis:sec}, and the text experiment protocol is given in Appendix~\ref{app:text}.

\bibliographystyle{iclr2027_conference}
\bibliography{references}

@misc{agarwal2023policy,
  title = {{On-Policy Distillation of Language Models: Learning from Self-Generated Mistakes}},
  author = {Rishabh Agarwal and Nino Vieillard and Zhou, Yongchao and Piotr Stańczyk and Sabela Ramos and Matthieu Geist and Olivier Bachem},
  year = {2023},
  doi = {10.48550/arxiv.2306.13649},
  journal = {{arXiv (Cornell University)}},
  url = {http://arxiv.org/abs/2306.13649},
}

@misc{an2025pard,
  title = {{PARD: Accelerating LLM Inference with Low-Cost PARallel Draft Model Adaptation}},
  author = {Zihao An and Huajun Bai and Ziqiong Liu and Dong Li and Emad Barsoum},
  year = {2025},
  doi = {10.48550/arxiv.2504.18583},
  eprint = {2504.18583},
  archivePrefix = {arXiv},
  journal = {{arXiv.org}},
  url = {https://arxiv.org/abs/2504.18583},
}

@misc{an2026pard2,
  title = {{PARD-2: Target-Aligned Parallel Draft Model for Dual-Mode Speculative Decoding}},
  author = {Zihao An and Taichi Liu and Ziqiong Liu and Dong Li and Ruofeng Liu and Emad Barsoum},
  year = {2026},
  doi = {10.48550/arxiv.2605.08632},
  eprint = {2605.08632},
  archivePrefix = {arXiv},
  journal = {{arXiv.org}},
  url = {https://arxiv.org/abs/2605.08632},
}

@misc{bai2025qwen3vl,
  title = {{Qwen3-VL Technical Report}},
  author = {Shuai Bai and Yuxuan Cai and Ruizhe Chen and Keqin Chen and Xionghui Chen and Zesen Cheng and Lianghao Deng and Wei Ding and Chang Gao and Chunjiang Ge and Wenbin Ge and Zhifang Guo and Qidong Huang and Jie Huang and Fei Huang and Binyuan Hui and Shutong Jiang and Zhaohai Li and Mingsheng Li and Mei Li and Kaixin Li and Zicheng Lin and Junyang Lin and Xuejing Liu and Jiawei Liu and Chenglong Liu and Yang Liu and Dayiheng Liu and Shixuan Liu and Dunjie Lu and Ruilin Luo and Chenxu Lv and Rui Men and Lingchen Meng and Xuancheng Ren and Xingzhang Ren and Sibo Song and Yuchong Sun and Jun Tang and Jianhong Tu and Jianqiang Wan and Peng Wang and Pengfei Wang and Qiuyue Wang and Yuxuan Wang and Tianbao Xie and Yiheng Xu and Haiyang Xu and Jin Xu and Zhibo Yang and Mingkun Yang and Jianxin Yang and An Yang and Bowen Yu and Fei Zhang and Hang Zhang and Xi Zhang and Bo Zheng and Humen Zhong and Jingren Zhou and Fan Zhou and Jing Zhou and Yuanzhi Zhu and Ke Zhu},
  year = {2025},
  doi = {10.48550/arxiv.2511.21631},
  eprint = {2511.21631},
  archivePrefix = {arXiv},
  journal = {{arXiv.org}},
  url = {https://arxiv.org/abs/2511.21631},
}

@misc{cai2024medusa,
  title = {{Medusa: Simple LLM Inference Acceleration Framework with Multiple Decoding Heads}},
  author = {Tianle Cai and Yuhong Li and Zhengyang Geng and Hongwu Peng and Jason D. Lee and Deming Chen and Tri Dao},
  year = {2024},
  doi = {10.48550/arxiv.2401.10774},
  journal = {{arXiv (Cornell University)}},
  url = {http://arxiv.org/abs/2401.10774},
}

@misc{chen2023accelerating,
  title = {{Accelerating Large Language Model Decoding with Speculative Sampling}},
  author = {Charlie Chen and Sebastian Borgeaud and Geoffrey Irving and Jean-Baptiste Lespiau and Laurent Sifre and John Jumper},
  year = {2023},
  doi = {10.48550/arxiv.2302.01318},
  journal = {{arXiv (Cornell University)}},
  url = {http://arxiv.org/abs/2302.01318},
}

@misc{chen2023sharegpt4v,
  title = {{ShareGPT4V: Improving Large Multi-modal Models with Better Captions}},
  author = {Lin Chen and Jinsong Li and Xiaoyi Dong and Pan Zhang and Conghui He and Jiaqi Wang and Feng Zhao and Dahua Lin},
  year = {2024},
  doi = {10.1007/978-3-031-72643-9_22},
  journal = {{Lecture notes in computer science}},
  url = {https://doi.org/10.1007/978-3-031-72643-9_22},
}

@misc{chen2024allava,
  title = {{ALLaVA: Harnessing GPT4V-Synthesized Data for Lite Vision-Language Models}},
  author = {Guiming Hardy Chen and Shunian Chen and Ruifei Zhang and Junying Chen and Xiangbo Wu and Zhiyi Zhang and Zhihong Chen and Jianquan Li and Xiang Wan and Benyou Wang},
  year = {2024},
  doi = {10.48550/arxiv.2402.11684},
  journal = {{arXiv (Cornell University)}},
  url = {http://arxiv.org/abs/2402.11684},
}

@misc{chen2026dflash,
  title = {{DFlash: Block Diffusion for Flash Speculative Decoding}},
  author = {Jian Chen and Yesheng Liang and Zhijian Liu},
  year = {2026},
  doi = {10.48550/arxiv.2602.06036},
  eprint = {2602.06036},
  archivePrefix = {arXiv},
  journal = {{arXiv.org}},
  url = {https://arxiv.org/abs/2602.06036},
}

@misc{ganesan2025massv,
  title = {{MASSV: Multimodal Adaptation and Self-Data Distillation for Speculative Decoding of Vision-Language Models}},
  author = {Mugilan Ganesan and Shane Segal and Ankur Aggarwal and Nish Sinnadurai and Sean Lie and Vithursan Thangarasa},
  year = {2025},
  doi = {10.18653/v1/2025.findings-emnlp.656},
  journal = {{Findings of the Association for Computational Linguistics: EMNLP 2025}},
  url = {https://doi.org/10.18653/v1/2025.findings-emnlp.656},
}

@misc{gao2025falcon,
  title = {{Falcon: Faster and Parallel Inference of Large Language Models Through Enhanced Semi-Autoregressive Drafting and Custom-Designed Decoding Tree}},
  author = {Xiangxiang Gao and Weisheng Xie and Yiwei Xiang and Feng Ji},
  year = {2025},
  doi = {10.1609/aaai.v39i22.34566},
  journal = {{Proceedings of the AAAI Conference on Artificial Intelligence}},
  url = {https://doi.org/10.1609/aaai.v39i22.34566},
}

@misc{goel2024direct,
  title = {{Direct Alignment of Draft Model for Speculative Decoding with Chat-Fine-Tuned LLMs}},
  author = {Raghavv Goel and Mukul Gagrani and Wonseok Jeon and Junyoung Park and Mingu Lee and Christopher Lott},
  year = {2024},
  doi = {10.48550/arxiv.2403.00858},
  journal = {{arXiv (Cornell University)}},
  url = {http://arxiv.org/abs/2403.00858},
}

@misc{gu2023minillm,
  title = {{MiniLLM: On-Policy Distillation of Large Language Models}},
  author = {Yuxian Gu and Li Dong and Furu Wei and Minlie Huang},
  year = {2023},
  doi = {10.48550/arxiv.2306.08543},
  journal = {{arXiv (Cornell University)}},
  url = {http://arxiv.org/abs/2306.08543},
}

@misc{hinton2015distilling,
  title = {{Distilling the Knowledge in a Neural Network}},
  author = {Geoffrey E. Hinton and O. Vinyals and J. Dean},
  year = {2015},
  eprint = {1503.02531},
  archivePrefix = {arXiv},
  journal = {{arXiv.org}},
  url = {https://arxiv.org/abs/1503.02531},
}

@misc{hong2025training,
  title = {{Training Domain Draft Models for Speculative Decoding: Best Practices and Insights}},
  author = {Hong, Fenglu and Ravi Raju and Jonathan Lingjie Li and Bo Li and Urmish Thakker and Avinash Ravichandran and Swayambhoo Jain and Changran Hu},
  year = {2025},
  doi = {10.48550/arxiv.2503.07807},
  journal = {{arXiv (Cornell University)}},
  url = {http://arxiv.org/abs/2503.07807},
}

@misc{hu2025dream,
  title = {{DREAM: Drafting with Refined Target Features and Entropy-Adaptive Cross-Attention Fusion for Multimodal Speculative Decoding}},
  author = {Yunhai Hu and Tianhua Xia and Zining Liu and Rahul Raman and Xingyu Liu and BO BAO and Eric Sather and Vithursan Thangarasa and Sai Qian Zhang},
  year = {2025},
  doi = {10.52202/085713-5586},
  journal = {{Advances in Neural Information Processing Systems 38}},
  url = {https://doi.org/10.52202/085713-5586},
}

@misc{kang2025vispec,
  title = {{ViSpec: Accelerating Vision-Language Models with Vision-Aware Speculative Decoding}},
  author = {Jialiang Kang and Han Shu and Wenshuo Li and Yingjie Zhai and Xinghao Chen},
  year = {2025},
  doi = {10.52202/085713-3852},
  journal = {{Advances in Neural Information Processing Systems 38}},
  url = {https://doi.org/10.52202/085713-3852},
}

@misc{kim2016sequence,
  title = {{Sequence-Level Knowledge Distillation}},
  author = {Yoon Kim and Alexander M. Rush},
  year = {2016},
  doi = {10.18653/v1/d16-1139},
  journal = {{Proceedings of the 2016 Conference on Empirical Methods in Natural
          Language Processing}},
  url = {https://doi.org/10.18653/v1/d16-1139},
}

@inproceedings{leviathan2022fast,
  title = {{Fast Inference from Transformers via Speculative Decoding}},
  author = {Yaniv Leviathan and Matan Kalman and Yossi Matias},
  booktitle = {Proceedings of the 40th International Conference on Machine Learning},
  year = {2023},
  volume = {202},
  series = {Proceedings of Machine Learning Research},
  pages = {19274--19286},
  publisher = {PMLR},
  url = {https://proceedings.mlr.press/v202/leviathan23a.html},
}

@misc{li2024eagle,
  title = {{EAGLE: Speculative Sampling Requires Rethinking Feature Uncertainty}},
  author = {Yuhui Li and Fangyun Wei and Chao Zhang and Hongyang Zhang},
  year = {2024},
  doi = {10.48550/arxiv.2401.15077},
  journal = {{arXiv (Cornell University)}},
  url = {http://arxiv.org/abs/2401.15077},
}

@misc{li2024eagle2,
  title = {{EAGLE-2: Faster Inference of Language Models with Dynamic Draft Trees}},
  author = {Yuhui Li and Fangyun Wei and Chao Zhang and Hongyang Zhang},
  year = {2024},
  doi = {10.18653/v1/2024.emnlp-main.422},
  url = {https://doi.org/10.18653/v1/2024.emnlp-main.422},
}

@misc{li2025eagle3,
  title = {{EAGLE-3: Scaling up Inference Acceleration of Large Language Models via Training-Time Test}},
  author = {Yuhui Li and Fangyun Wei and Chao Zhang and Hongyang Zhang},
  year = {2025},
  doi = {10.52202/085713-4562},
  url = {https://doi.org/10.52202/085713-4562},
}

@misc{lin2014coco,
  title = {{Microsoft COCO: Common Objects in Context}},
  author = {Tsung-Yi Lin and Michael Maire and Serge Belongie and James Hays and Pietro Perona and Deva Ramanan and Piotr Dollár and C. Lawrence Zitnick},
  year = {2014},
  doi = {10.1007/978-3-319-10602-1_48},
  journal = {{Lecture Notes in Computer Science}},
  url = {https://doi.org/10.1007/978-3-319-10602-1_48},
}

@misc{lin2025speculative,
  title = {{Speculative Decoding Reimagined for Multimodal Large Language Models}},
  author = {Lin, Luxi and Lin, Zhihang and Zhanpeng Zeng and Rongrong Ji},
  year = {2025},
  doi = {10.48550/arxiv.2505.14260},
  journal = {{arXiv (Cornell University)}},
  url = {http://arxiv.org/abs/2505.14260},
}

@misc{liu2023online,
  title = {{Online Speculative Decoding}},
  author = {Xiaoxuan Liu and Lanxiang Hu and Peter Bailis and Alvin Cheung and Zhijie Deng and Ion Stoica and Hao Zhang},
  year = {2023},
  doi = {10.48550/arxiv.2310.07177},
  journal = {{arXiv (Cornell University)}},
  url = {http://arxiv.org/abs/2310.07177},
}

@misc{liu2023visual,
  title = {{Visual Instruction Tuning}},
  author = {Haotian Liu and Chunyuan Li and Qingyang Wu and Yong Jae Lee},
  year = {2023},
  doi = {10.52202/075280-1516},
  journal = {{Advances in Neural Information Processing Systems 36}},
  url = {https://doi.org/10.52202/075280-1516},
}

@misc{mathew2021docvqa,
  title = {{DocVQA: A Dataset for VQA on Document Images}},
  author = {Minesh Mathew and Dimosthenis Karatzas and C. V. Jawahar},
  year = {2021},
  doi = {10.1109/wacv48630.2021.00225},
  journal = {{2021 IEEE Winter Conference on Applications of Computer Vision (WACV)}},
  url = {https://doi.org/10.1109/wacv48630.2021.00225},
}

@misc{ross2010reduction,
  title = {{A Reduction of Imitation Learning and Structured Prediction to No-Regret Online Learning}},
  author = {Stéphane Ross and Geoffrey J. Gordon and J. Andrew Bagnell},
  year = {2010},
  doi = {10.48550/arXiv.1011.0686},
  journal = {{arXiv (Cornell University)}},
  url = {https://arxiv.org/abs/1011.0686},
}

@misc{singh2019textvqa,
  title = {{Towards VQA Models That Can Read}},
  author = {Amanpreet Singh and Vivek Natarajan and Meet Shah and Yu Jiang and Xinlei Chen and Dhruv Batra and Devi Parikh and Marcus Rohrbach},
  year = {2019},
  doi = {10.1109/cvpr.2019.00851},
  eprint = {1904.08920},
  archivePrefix = {arXiv},
  journal = {{Computer Vision and Pattern Recognition}},
  url = {https://arxiv.org/abs/1904.08920},
}

@misc{stern2018blockwise,
  title = {{Blockwise Parallel Decoding for Deep Autoregressive Models}},
  author = {Mitchell Stern and Noam Shazeer and Jakob Uszkoreit},
  year = {2018},
  doi = {10.48550/arxiv.1811.03115},
  journal = {{arXiv (Cornell University)}},
  url = {http://arxiv.org/abs/1811.03115},
}

@misc{xiao2024parallelspec,
  title = {{ParallelSpec: Parallel Drafter for Efficient Speculative Decoding}},
  author = {Zilin Xiao and Hongming Zhang and Tao Ge and Ouyang, Siru and Vicente Ordóñez and Dong Yu},
  year = {2024},
  doi = {10.48550/arxiv.2410.05589},
  journal = {{arXiv (Cornell University)}},
  url = {http://arxiv.org/abs/2410.05589},
}

@misc{zafrir2025fastdraft,
  title = {{FastDraft: How to Train Your Draft}},
  author = {Ofir Zafrir and Igor Margulis and Dorin Shteyman and Shira Guskin and Guy Boudoukh},
  year = {2025},
  doi = {10.18653/v1/2025.findings-acl.1156},
  journal = {{Findings of the Association for Computational Linguistics: ACL 2025}},
  url = {https://doi.org/10.18653/v1/2025.findings-acl.1156},
}

@misc{zhang2024learning,
  title = {{Learning Harmonized Representations for Speculative Sampling}},
  author = {Lefan Zhang and Xiaodan Wang and Yanhua Huang and Ruiwen Xu},
  year = {2024},
  doi = {10.48550/arxiv.2408.15766},
  journal = {{arXiv (Cornell University)}},
  url = {http://arxiv.org/abs/2408.15766},
}

@misc{zhou2023distillspec,
  title = {{DistillSpec: Improving Speculative Decoding via Knowledge Distillation}},
  author = {Yongchao Zhou and Kaifeng Lyu and Ankit Singh Rawat and Aditya Krishna Menon and Afshin Rostamizadeh and Sanjiv Kumar and Jean-François Kagy and Rishabh Agarwal},
  year = {2023},
  doi = {10.48550/arxiv.2310.08461},
  journal = {{arXiv (Cornell University)}},
  url = {http://arxiv.org/abs/2310.08461},
}

@misc{lee2026answerconditioned,
  title = {{Answer-Conditioned Chains of Thought Degrade Verifiable-Reasoning Distillation in Large Language Models}},
  author = {Jungseob Lee and Seungyoon Lee and Suhyune Son and Dongyub Jude Lee and Sungbin Han and Sugyeong Eo and Heuiseok Lim},
  year = {2026},
  eprint = {2607.14552},
  archivePrefix = {arXiv},
  journal = {{arXiv preprint arXiv:2607.14552}},
  url = {https://arxiv.org/abs/2607.14552},
}

@misc{tong2026sage,
  title = {{SAGE: Accelerating Vision-Language Models via Entropy-Guided Adaptive Speculative Decoding}},
  author = {Yujia Tong and Tian Zhang and Yunyang Wan and Kaiwei Lin and Jingling Yuan and Chuang Hu},
  year = {2026},
  eprint = {2602.00523},
  archivePrefix = {arXiv},
  primaryClass = {cs.CV},
  doi = {10.48550/arXiv.2602.00523},
  url = {https://arxiv.org/abs/2602.00523},
}

@misc{yang2025qwen3,
  title = {{Qwen3 Technical Report}},
  author = {An Yang and Anfeng Li and Baosong Yang and Beichen Zhang and Binyuan Hui and Bo Zheng and Bowen Yu and Chang Gao and Chengen Huang and Chenxu Lv and Chujie Zheng and Dayiheng Liu and Fan Zhou and Fei Huang and Feng Hu and Hao Ge and Haoran Wei and Huan Lin and Jialong Tang and Jian Yang and Jianhong Tu and Jianwei Zhang and Jianxin Yang and Jiaxi Yang and Jing Zhou and Jingren Zhou and Junyang Lin and Kai Dang and Keqin Bao and Kexin Yang and Le Yu and Lianghao Deng and Mei Li and Mingfeng Xue and Mingze Li and Pei Zhang and Peng Wang and Qin Zhu and Rui Men and Ruize Gao and Shixuan Liu and Shuang Luo and Tianhao Li and Tianyi Tang and Wenbiao Yin and Xingzhang Ren and Xinyu Wang and Xinyu Zhang and Xuancheng Ren and Yang Fan and Yang Su and Yichang Zhang and Yinger Zhang and Yu Wan and Yuqiong Liu and Zekun Wang and Zeyu Cui and Zhenru Zhang and Zhipeng Zhou and Zihan Qiu},
  year = {2025},
  doi = {10.48550/arxiv.2505.09388},
  eprint = {2505.09388},
  archivePrefix = {arXiv},
  journal = {{arXiv preprint}},
  url = {https://arxiv.org/abs/2505.09388},
}

@inproceedings{ding2023ultrachat,
    title = "Enhancing Chat Language Models by Scaling High-quality Instructional Conversations",
    author = "Ding, Ning  and
      Chen, Yulin  and
      Xu, Bokai  and
      Qin, Yujia  and
      Hu, Shengding  and
      Liu, Zhiyuan  and
      Sun, Maosong  and
      Zhou, Bowen",
    editor = "Bouamor, Houda  and
      Pino, Juan  and
      Bali, Kalika",
    booktitle = "Proceedings of the 2023 Conference on Empirical Methods in Natural Language Processing",
    month = dec,
    year = "2023",
    address = "Singapore",
    publisher = "Association for Computational Linguistics",
    url = "https://aclanthology.org/2023.emnlp-main.183/",
    doi = "10.18653/v1/2023.emnlp-main.183",
    pages = "3029--3051"
}

@inproceedings{zheng2023judging,
 author = {Zheng, Lianmin and Chiang, Wei-Lin and Sheng, Ying and Zhuang, Siyuan and Wu, Zhanghao and Zhuang, Yonghao and Lin, Zi and Li, Zhuohan and Li, Dacheng and Xing, Eric and Zhang, Hao and Gonzalez, Joseph and Stoica, Ion},
 booktitle = {Advances in Neural Information Processing Systems},
 doi = {10.52202/075280-2020},
 editor = {A. Oh and T. Naumann and A. Globerson and K. Saenko and M. Hardt and S. Levine},
 pages = {46595--46623},
 publisher = {Curran Associates, Inc.},
 title = {Judging LLM-as-a-Judge with MT-Bench and Chatbot Arena},
 url = {https://proceedings.neurips.cc/paper_files/paper/2023/file/91f18a1287b398d378ef22505bf41832-Paper-Datasets_and_Benchmarks.pdf},
 volume = {36},
 year = {2023}
}

@misc{cobbe2021training,
  title = {{Training Verifiers to Solve Math Word Problems}},
  author = {Karl Cobbe and Vineet Kosaraju and Mohammad Bavarian and Mark Chen and Heewoo Jun and Lukasz Kaiser and Matthias Plappert and Jerry Tworek and Jacob Hilton and Reiichiro Nakano and Christopher Hesse and John Schulman},
  year = {2021},
  doi = {10.48550/arxiv.2110.14168},
  eprint = {2110.14168},
  archivePrefix = {arXiv},
  journal = {{arXiv preprint}},
  url = {https://arxiv.org/abs/2110.14168},
}

@misc{chen2021evaluating,
  title = {{Evaluating Large Language Models Trained on Code}},
  author = {Mark Chen and Jerry Tworek and Heewoo Jun and Qiming Yuan and Henrique Ponde de Oliveira Pinto and Jared Kaplan and Harri Edwards and Yuri Burda and Nicholas Joseph and Greg Brockman and Alex Ray and Raul Puri and Gretchen Krueger and Michael Petrov and Heidy Khlaaf and Girish Sastry and Pamela Mishkin and Brooke Chan and Scott Gray and Nick Ryder and Mikhail Pavlov and Alethea Power and Lukasz Kaiser and Mohammad Bavarian and Clemens Winter and Philippe Tillet and Felipe Petroski Such and Dave Cummings and Matthias Plappert and Fotios Chantzis and Elizabeth Barnes and Ariel Herbert-Voss and William Hebgen Guss and Alex Nichol and Alex Paino and Nikolas Tezak and Jie Tang and Igor Babuschkin and Suchir Balaji and Shantanu Jain and William Saunders and Christopher Hesse and Andrew N. Carr and Jan Leike and Josh Achiam and Vedant Misra and Evan Morikawa and Alec Radford and Matthew Knight and Miles Brundage and Mira Murati and Katie Mayer and Peter Welinder and Bob McGrew and Dario Amodei and Sam McCandlish and Ilya Sutskever and Wojciech Zaremba},
  year = {2021},
  doi = {10.48550/arxiv.2107.03374},
  eprint = {2107.03374},
  archivePrefix = {arXiv},
  journal = {{arXiv preprint}},
  url = {https://arxiv.org/abs/2107.03374},
}

@misc{jain2024livecodebench,
  title = {{LiveCodeBench: Holistic and Contamination Free Evaluation of Large Language Models for Code}},
  author = {Naman Jain and King Han and Alex Gu and Wen-Ding Li and Fanjia Yan and Tianjun Zhang and Sida Wang and Armando Solar-Lezama and Koushik Sen and Ion Stoica},
  year = {2024},
  doi = {10.48550/arxiv.2403.07974},
  eprint = {2403.07974},
  archivePrefix = {arXiv},
  journal = {{arXiv preprint}},
  url = {https://arxiv.org/abs/2403.07974},
}

@misc{zhang2025aime25,
  title = {{American Invitational Mathematics Examination (AIME) 2025}},
  author = {Yifan Zhang and {Team Math-AI}},
  year = {2025},
  howpublished = {Math-AI dataset release on Hugging Face},
  url = {https://math-ai-org.github.io/aime25/},
}

@misc{lee2022norose,
  title = {{There is no rose without a thorn: Finding weaknesses on BlenderBot 2.0 in terms of Model, Data and User-Centric Approach}},
  author = {Jungseob Lee and Midan Shim and Suhyune Son and Chanjun Park and Yujin Kim and Heuiseok Lim},
  year = {2022},
  doi = {10.48550/arXiv.2201.03239},
  eprint = {2201.03239},
  archivePrefix = {arXiv},
  journal = {arXiv preprint},
  url = {https://arxiv.org/abs/2201.03239},
}

\clearpage
\appendix

\section{Training Configuration}
\label{app:training}

Table~\ref{app:tab-train} lists the shared vision-language training configuration. These arms differ only in the objective, number of epochs, rollout depth $R$ or envelope $\gamma$ named in the comparison. Appendix~\ref{app:text} gives the text-only configuration.

\begin{table}[ht]
\centering
\caption{Shared vision-language training configuration.}
\label{app:tab-train}
\small
\begin{tabular}{@{}p{0.34\linewidth}p{0.6\linewidth}@{}}
\toprule
Setting & Value \\
\midrule
\grouprow{2}{Target} \\
Model & Qwen3-VL-8B-Instruct \\
Second target & Qwen3-VL-4B-Instruct, text decoder hidden 2560, 36 layers, tied embeddings \\
Text decoder hidden size & 4096 \\
Text decoder layers & 36 \\
Attention heads & 32 \\
Vocabulary & 151,936 \\
\grouprow{2}{Drafter} \\
Architecture & DFlash block drafter, 5 layers \\
Target feature layers & 1, 9, 17, 25, 33 \\
Block & 16 slots, slot 0 the anchor, slots 1 to 15 predicted in parallel \\
Warm start & \texttt{z-lab/Qwen3-8B-DFlash-b16}, text-only; \texttt{z-lab/Qwen3-4B-DFlash-b16} for the 4B target \\
Architecture after warm start & kept unchanged \\
\grouprow{2}{Data} \\
ALLaVA dataset & \texttt{FreedomIntelligence/ALLaVA-4V}, instruction file of the \texttt{allava\_laion} subset \\
ALLaVA content & LAION images, GPT-4V-style responses \\
ALLaVA selection & rows shuffled with a fixed seed, then the first 26,196 that are single-turn, have their image in the first released image chunk and a response of at least 40 characters \\
ShareGPT4V source & GPT-4V captions of \texttt{cap100k} whose images come from LLaVA-Pretrain (LAION, CC, SBU), 30,000 rows \\
ShareGPT4V selection & sorted by id, shuffled with a fixed seed, first 26,196 kept to match the ALLaVA row count \\
Maximum length & 4096 tokens \\
Image size & 50,176 to 200,704 pixels \\
Vision tokens per image & 16 to 64 after merging \\
\grouprow{2}{Optimisation} \\
Hardware & 4 H100 80\,GB GPUs, FSDP \\
Batch size & 6 per GPU \\
Learning rate & $1.4697\times10^{-3}$, the same for every objective \\
Optimiser & AdamW on fp32 copies of the bf16 weights, $\beta=(0.9, 0.999)$, no weight decay \\
Schedule & cosine annealing after a linear warmup over 4\% of the steps \\
Gradient-norm clip & 1.0 \\
Epochs & 3, unless a row names another schedule \\
Steps for 3 epochs, ALLaVA & 3,261 \\
Steps for 3 epochs, ShareGPT4V & 3,273 \\
Anchors per sample & 128 \\
Precision & bf16 \\
Attention & sdpa \\
\grouprow{2}{Loss} \\
KD loss & soft cross-entropy against the target's top-8 probabilities plus one bucket for the remaining mass, weighted mean over slots \\
KD loss scale $\lambda$ & 1.013441, matching the cross-entropy loss scale at initialisation \\
Slot envelope & $w_k=e^{-(k-1)/\gamma}$ \\
Envelope $\gamma$ & 2, except one erase schedule at 0.5 \\
Rollout depth $R$ & 14 for ALR and ALR + IRA \\
Depth sweep, ALR alone & $R\in\{2,4,6\}$ \\
Secondary-anchor offset & $j\sim\mathcal{U}\{2,\dots,14\}$ \\
\bottomrule
\end{tabular}
\end{table}

\paragraph{Rollout implementation.}
The rollout runs inside the training step and reuses the target's forward pass over the corpus sample. That pass fills a key-value cache, and the rollout then takes $R$ greedy steps, each of which runs all anchors of a sample through the target's decoder layers as one batch. The query of an anchor attends under a single softmax to the cached corpus keys up to its anchor, which all anchors share, and to its own rollout keys so far, which are kept in a separate per-anchor buffer. The corpus keys are therefore never copied per anchor, and grouped-query attention forms its scores per key-value head without repeating them. Rollout positions continue the anchor's text position, with the three M-RoPE axes advancing together. Each step reads the labels from the target's final normalised hidden state through its output head, keeping the 8 most probable tokens and the remaining mass, and greedy ties go to the lower token id. For IRA, the same steps also store the outputs of the five target layers the drafter reads for $y^*_1,\dots,y^*_{13}$, which become the context of the secondary blocks. On four H100 GPUs, an epoch on the ALLaVA rows takes 0.75 GPU-hours for erase, 1.49 for ALR, 0.95 for ALR at $R{=}4$ and 1.32 for ALR + IRA.

\section{Evaluation Protocol}
\label{app:eval}

Table~\ref{app:tab-eval} lists the DFlash-family vision-language evaluation protocol. The images of both training corpora come from LAION, CC and SBU, and none of the evaluation image sources is shared with either training corpus.

\begin{table}[ht]
\centering
\caption{DFlash-family vision-language evaluation protocol.}
\label{app:tab-eval}
\small
\begin{tabular}{@{}p{0.34\linewidth}p{0.6\linewidth}@{}}
\toprule
Setting & Value \\
\midrule
\grouprow{2}{Prompts} \\
Captioning & COCO val2017, detailed captioning \\
Captioning prompt & ``Describe this image in detail.'' \\
TextVQA & training split \\
DocVQA & validation split \\
Prompts per domain & the first 100, of which 99 are scored after one warm-up \\
Maximum new tokens & 256 \\
Temperature & 0, and 1 for the sampling readings \\
\grouprow{2}{Decoder} \\
Verification & tree, one target forward pass per round \\
Tree budget & 63 drafted tokens \\
Candidates per slot & the drafter's 8 most probable tokens \\
Tree construction & best-first by path log-probability, prefix-closed; a path can reach all 15 slots \\
Mean acceptance length $\tau$ (MAT) & accepted draft tokens plus the one token the target adds in each verification round \\
Chain decoding & one robustness reading: the top token of all 15 slots, verified as one chain \\
Attention & sdpa \\
Precision & bf16 \\
\grouprow{2}{Hardware} \\
Batch size & one prompt at a time, for both the drafter and the baseline \\
MAT & one RTX A6000 per evaluation; one H100 for the 4B target \\
Speed & one H100 assigned to each evaluator \\
Speed baseline & autoregressive decoding timed in the same run \\
\bottomrule
\end{tabular}
\end{table}

\section{Vision-Language Results by Domain}
\label{app:vision-results}

Table~\ref{app:vision-domains} disaggregates the vision-language results of Table~\ref{results:tab-mat} by evaluation domain and reports training-seed variation. ALLaVA and ShareGPT4V name training corpora, while COCO, TextVQA and DocVQA name evaluation datasets.

\begin{table}[t]
\centering
\caption{Vision-language mean acceptance length $\tau$ by evaluation domain after three training epochs. Subscripts show training-seed standard deviations. Avg. is the geometric mean across domains within each seed, averaged across seeds.}
\label{app:vision-domains}
\small
\setlength{\tabcolsep}{2pt}
\begin{tabularx}{\linewidth}{l*{4}{>{\centering\arraybackslash}X}}
\toprule
Method & COCO & TextVQA & DocVQA & Avg. \\
\midrule
\grouprow{5}{Qwen3-VL-8B trained on ALLaVA, $T{=}0$} \\
DFlash & $2.563_{\pm0.006}$ & $2.655_{\pm0.010}$ & $3.501_{\pm0.010}$ & $2.878_{\pm0.007}$ \\
Erase & $3.124_{\pm0.006}$ & $3.375_{\pm0.005}$ & $4.204_{\pm0.037}$ & $3.539_{\pm0.010}$ \\
ALR & $3.242_{\pm0.007}$ & $3.655_{\pm0.024}$ & $4.261_{\pm0.016}$ & $3.696_{\pm0.005}$ \\
\rowcolor{ourtint}
ALR + IRA & $3.376_{\pm0.007}$ & $3.851_{\pm0.011}$ & $4.361_{\pm0.065}$ & $3.841_{\pm0.021}$ \\
\midrule
\grouprow{5}{Qwen3-VL-8B trained on ALLaVA, $T{=}1$} \\
DFlash & $2.435_{\pm0.009}$ & $2.601_{\pm0.013}$ & $3.436_{\pm0.005}$ & $2.792_{\pm0.002}$ \\
Erase & $2.952_{\pm0.007}$ & $3.196_{\pm0.012}$ & $4.115_{\pm0.021}$ & $3.386_{\pm0.009}$ \\
ALR & $3.015_{\pm0.012}$ & $3.422_{\pm0.024}$ & $4.140_{\pm0.026}$ & $3.495_{\pm0.007}$ \\
\rowcolor{ourtint}
ALR + IRA & $3.150_{\pm0.006}$ & $3.615_{\pm0.015}$ & $4.261_{\pm0.026}$ & $3.647_{\pm0.005}$ \\
\midrule
\grouprow{5}{Qwen3-VL-8B trained on ShareGPT4V, $T{=}0$} \\
DFlash & $2.542_{\pm0.009}$ & $2.691_{\pm0.004}$ & $3.216_{\pm0.028}$ & $2.802_{\pm0.003}$ \\
Erase & $3.403_{\pm0.010}$ & $3.326_{\pm0.023}$ & $4.023_{\pm0.036}$ & $3.571_{\pm0.016}$ \\
ALR & $3.614_{\pm0.003}$ & $3.398_{\pm0.009}$ & $4.035_{\pm0.017}$ & $3.673_{\pm0.007}$ \\
\rowcolor{ourtint}
ALR + IRA & $3.817_{\pm0.024}$ & $3.600_{\pm0.024}$ & $4.073_{\pm0.040}$ & $3.825_{\pm0.029}$ \\
\midrule
\grouprow{5}{Qwen3-VL-8B trained on ShareGPT4V, $T{=}1$} \\
DFlash & $2.395_{\pm0.003}$ & $2.599_{\pm0.007}$ & $3.183_{\pm0.000}$ & $2.706_{\pm0.003}$ \\
Erase & $3.132_{\pm0.000}$ & $3.207_{\pm0.031}$ & $3.874_{\pm0.014}$ & $3.388_{\pm0.015}$ \\
ALR & $3.306_{\pm0.020}$ & $3.249_{\pm0.004}$ & $3.910_{\pm0.021}$ & $3.476_{\pm0.015}$ \\
\rowcolor{ourtint}
ALR + IRA & $3.478_{\pm0.004}$ & $3.379_{\pm0.019}$ & $3.975_{\pm0.030}$ & $3.602_{\pm0.017}$ \\
\midrule
\grouprow{5}{Qwen3-VL-4B trained on ALLaVA, $T{=}0$} \\
DFlash & $2.619_{\pm0.005}$ & $2.704_{\pm0.008}$ & $3.529_{\pm0.003}$ & $2.924_{\pm0.006}$ \\
Erase & $3.185_{\pm0.007}$ & $3.347_{\pm0.016}$ & $4.215_{\pm0.044}$ & $3.555_{\pm0.004}$ \\
ALR & $3.259_{\pm0.002}$ & $3.550_{\pm0.025}$ & $4.124_{\pm0.051}$ & $3.627_{\pm0.024}$ \\
\rowcolor{ourtint}
ALR + IRA & $3.392_{\pm0.002}$ & $3.785_{\pm0.017}$ & $4.303_{\pm0.019}$ & $3.809_{\pm0.011}$ \\
\midrule
\grouprow{5}{Qwen3-VL-4B trained on ALLaVA, $T{=}1$} \\
DFlash & $2.476_{\pm0.000}$ & $2.554_{\pm0.001}$ & $3.383_{\pm0.007}$ & $2.776_{\pm0.002}$ \\
Erase & $2.947_{\pm0.004}$ & $3.105_{\pm0.024}$ & $4.046_{\pm0.014}$ & $3.333_{\pm0.011}$ \\
ALR & $3.010_{\pm0.011}$ & $3.287_{\pm0.042}$ & $4.008_{\pm0.047}$ & $3.410_{\pm0.024}$ \\
\rowcolor{ourtint}
ALR + IRA & $3.105_{\pm0.005}$ & $3.440_{\pm0.018}$ & $4.135_{\pm0.024}$ & $3.535_{\pm0.002}$ \\
\bottomrule
\end{tabularx}
\end{table}

\section{Text-Only Evaluation}
\label{app:text}

We train DFlash, erase, ALR and ALR + IRA for three epochs on the same fixed UltraChat subset with Qwen3-4B as the target and thinking disabled. The selection contains 26,196 conversations, of which 25,555 pass the minimum assistant-token filter. All methods start from \texttt{z-lab/Qwen3-4B-DFlash-b16} and use two training replicates, effective batch size 24, learning rate $1.4697\times10^{-3}$, maximum training length 1,024, 128 anchors, block size 16, $\gamma=2$ and KD scale 1.0. Corpus selection and data order stay fixed. The four-H100 training configuration uses batch size 6 per GPU.

Evaluation covers seven benchmarks at $T{=}0$ and $T{=}1$, with at most 256 generated tokens. It includes all 80 MT-Bench first turns \citep{zheng2023judging}, 1,319 GSM8K test questions \citep{cobbe2021training} and 164 HumanEval tasks \citep{chen2021evaluating}, plus the four benchmarks described below. The full-benchmark evaluations use one H100, bf16, SDPA and the public DFlash chain decoder. At $T{=}1$, target sampling preserves the target distribution in exact arithmetic but does not force identical outputs across methods. We measure decoding efficiency rather than answer accuracy or code pass rate. Acceptance length pools verification rounds within each task. The Text panel of Table~\ref{results:tab-mat} reports task means and their aggregate. Tables~\ref{results:text-benchmarks} and~\ref{app:text-extra} add standard deviations across the two training seeds. The aggregate first takes the geometric mean across all seven tasks within each seed, then averages across seeds.

Within each text benchmark and temperature, all methods use the same AR timing reference. Model loading and tokenization are excluded.

\begin{table}[t]
\centering
\caption{Text-only generation speedup over autoregressive decoding and mean accepted length $\tau$ on Qwen3-4B. All methods train for three epochs on the same fixed UltraChat corpus. Subscripts show standard deviations across training seeds.}
\label{results:text-benchmarks}
\small
\setlength{\tabcolsep}{2pt}
\begin{tabularx}{\linewidth}{l*{6}{>{\centering\arraybackslash}X}}
\toprule
& \multicolumn{2}{c}{MT-Bench} & \multicolumn{2}{c}{GSM8K} & \multicolumn{2}{c}{HumanEval} \\
\cmidrule(lr){2-3}\cmidrule(lr){4-5}\cmidrule(lr){6-7}
Method & Speedup & $\tau$ & Speedup & $\tau$ & Speedup & $\tau$ \\
\midrule
\grouprow{7}{$T{=}0$} \\
DFlash & $1.82_{\pm0.07}\!\times$ & $2.50_{\pm0.01}$ & $2.79_{\pm0.05}\!\times$ & $3.96_{\pm0.01}$ & $2.98_{\pm0.07}\!\times$ & $4.22_{\pm0.02}$ \\
Erase & $2.07_{\pm0.04}\!\times$ & $2.85_{\pm0.01}$ & $3.16_{\pm0.08}\!\times$ & $4.48_{\pm0.01}$ & $3.46_{\pm0.10}\!\times$ & $4.92_{\pm0.00}$ \\
ALR & $2.09_{\pm0.01}\!\times$ & $2.94_{\pm0.02}$ & $3.38_{\pm0.09}\!\times$ & $4.76_{\pm0.00}$ & $3.56_{\pm0.13}\!\times$ & $4.98_{\pm0.06}$ \\
\rowcolor{ourtint}
ALR + IRA & $2.13_{\pm0.02}\!\times$ & $2.99_{\pm0.00}$ & $3.55_{\pm0.06}\!\times$ & $4.83_{\pm0.07}$ & $3.63_{\pm0.00}\!\times$ & $5.02_{\pm0.06}$ \\
\midrule
\grouprow{7}{$T{=}1$} \\
DFlash & $1.79_{\pm0.03}\!\times$ & $2.42_{\pm0.05}$ & $2.71_{\pm0.05}\!\times$ & $3.78_{\pm0.01}$ & $2.89_{\pm0.02}\!\times$ & $3.97_{\pm0.07}$ \\
Erase & $1.97_{\pm0.04}\!\times$ & $2.72_{\pm0.04}$ & $3.01_{\pm0.09}\!\times$ & $4.26_{\pm0.01}$ & $3.22_{\pm0.02}\!\times$ & $4.57_{\pm0.10}$ \\
ALR & $2.06_{\pm0.03}\!\times$ & $2.79_{\pm0.02}$ & $3.19_{\pm0.08}\!\times$ & $4.50_{\pm0.03}$ & $3.19_{\pm0.07}\!\times$ & $4.63_{\pm0.00}$ \\
\rowcolor{ourtint}
ALR + IRA & $2.15_{\pm0.04}\!\times$ & $2.83_{\pm0.01}$ & $3.29_{\pm0.05}\!\times$ & $4.55_{\pm0.02}$ & $3.20_{\pm0.11}\!\times$ & $4.66_{\pm0.01}$ \\
\bottomrule
\end{tabularx}
\end{table}

\paragraph{Additional text benchmarks.}
The additional benchmarks comprise 80 Alpaca prompts distributed with EAGLE-3 \citep{li2025eagle3}, all 30 AIME24 problems from the 2024 I and II contests,\footnote{We use the math-ai/aime24 test split at the fixed revision \href{https://huggingface.co/datasets/math-ai/aime24/tree/83a7f387baaa524a8bda0022eac0541582297103}{\texttt{83a7f387}}.} 30 AIME25 problems \citep{zhang2025aime25} and 400 LiveCodeBench release\_v1 tasks \citep{jain2024livecodebench}. Dataset revisions and prompt formatting are fixed across all four methods and both training seeds. The target, decoder, temperatures and 256-token output limit remain the same. Table~\ref{app:text-extra} reports seed variation for these four benchmarks.

\begin{table}[t]
\centering
\caption{Generation speedup over autoregressive decoding and mean acceptance length $\tau$ on additional text benchmarks. Subscripts show standard deviations across two training seeds.}
\label{app:text-extra}
\small
\setlength{\tabcolsep}{2pt}
\begin{tabularx}{\linewidth}{l*{4}{>{\centering\arraybackslash}X}}
\toprule
 & \multicolumn{2}{c}{$T{=}0$} & \multicolumn{2}{c}{$T{=}1$} \\
\cmidrule(lr){2-3}\cmidrule(lr){4-5}
Method & Speedup & $\tau$ & Speedup & $\tau$ \\
\midrule
\grouprow{5}{Alpaca} \\
DFlash & $1.56_{\pm0.02}\!\times$ & $2.17_{\pm0.00}$ & $1.52_{\pm0.04}\!\times$ & $2.16_{\pm0.01}$ \\
Erase & $1.88_{\pm0.01}\!\times$ & $2.52_{\pm0.01}$ & $1.72_{\pm0.09}\!\times$ & $2.47_{\pm0.01}$ \\
ALR & $1.87_{\pm0.07}\!\times$ & $2.58_{\pm0.02}$ & $1.78_{\pm0.13}\!\times$ & $2.54_{\pm0.01}$ \\
\rowcolor{ourtint}
ALR + IRA & $1.92_{\pm0.04}\!\times$ & $2.62_{\pm0.00}$ & $1.80_{\pm0.08}\!\times$ & $2.57_{\pm0.00}$ \\
\midrule
\grouprow{5}{AIME24} \\
DFlash & $2.61_{\pm0.02}\!\times$ & $3.69_{\pm0.03}$ & $2.41_{\pm0.03}\!\times$ & $3.35_{\pm0.05}$ \\
Erase & $2.85_{\pm0.04}\!\times$ & $4.06_{\pm0.01}$ & $2.49_{\pm0.04}\!\times$ & $3.54_{\pm0.07}$ \\
ALR & $2.91_{\pm0.07}\!\times$ & $4.20_{\pm0.04}$ & $2.58_{\pm0.14}\!\times$ & $3.66_{\pm0.05}$ \\
\rowcolor{ourtint}
ALR + IRA & $2.97_{\pm0.11}\!\times$ & $4.27_{\pm0.05}$ & $2.59_{\pm0.02}\!\times$ & $3.65_{\pm0.04}$ \\
\midrule
\grouprow{5}{AIME25} \\
DFlash & $2.76_{\pm0.14}\!\times$ & $4.07_{\pm0.04}$ & $2.40_{\pm0.04}\!\times$ & $3.49_{\pm0.13}$ \\
Erase & $3.01_{\pm0.03}\!\times$ & $4.34_{\pm0.02}$ & $2.62_{\pm0.05}\!\times$ & $3.77_{\pm0.06}$ \\
ALR & $2.94_{\pm0.04}\!\times$ & $4.53_{\pm0.08}$ & $2.46_{\pm0.03}\!\times$ & $3.85_{\pm0.06}$ \\
\rowcolor{ourtint}
ALR + IRA & $3.14_{\pm0.01}\!\times$ & $4.66_{\pm0.06}$ & $2.66_{\pm0.04}\!\times$ & $3.96_{\pm0.01}$ \\
\midrule
\grouprow{5}{LiveCodeBench} \\
DFlash & $2.33_{\pm0.01}\!\times$ & $3.25_{\pm0.05}$ & $2.20_{\pm0.06}\!\times$ & $3.00_{\pm0.04}$ \\
Erase & $2.64_{\pm0.06}\!\times$ & $3.72_{\pm0.01}$ & $2.44_{\pm0.00}\!\times$ & $3.38_{\pm0.02}$ \\
ALR & $2.62_{\pm0.11}\!\times$ & $3.81_{\pm0.03}$ & $2.42_{\pm0.09}\!\times$ & $3.47_{\pm0.01}$ \\
\rowcolor{ourtint}
ALR + IRA & $2.73_{\pm0.13}\!\times$ & $3.84_{\pm0.02}$ & $2.52_{\pm0.15}\!\times$ & $3.49_{\pm0.01}$ \\
\bottomrule
\end{tabularx}
\end{table}

\paragraph{Repeated timing.}
Table~\ref{app:text-timing} tests the aggregate speed advantage with fresh AR references and balanced execution order. We select 32 prompts per benchmark by identifier hash and all 30 from each AIME set, giving 220 fixed prompts. The target, decoder and output limit remain unchanged. After warm-up, each prompt runs on the same A100 for all methods. For each prompt and temperature, two trained checkpoint pairs and three timing repeats cover all six AR/erase/ALR + IRA execution orders. Sampling randomness stays fixed across timing repeats. We average total generation milliseconds per token over repeats and then prompts within each task. Speedups are geometric means of task latency ratios within each checkpoint pair, averaged over pairs. Gain uses the erase/ALR + IRA ratio directly. The 95\% intervals use 10,000 paired prompt resamples stratified by task, retaining all methods, checkpoints, temperatures and repeats together. They condition on the measured hardware and trained checkpoint pairs.

\begin{table}[t]
\centering
\caption{Aggregate generation speedup over AR on 220 fixed prompts from seven text benchmarks, measured on A100. Gain compares ALR + IRA with erase. Brackets give paired-prompt 95\% bootstrap intervals.}
\label{app:text-timing}
\small
\setlength{\tabcolsep}{8pt}
\begin{tabular}{lcc}
\toprule
Method & $T{=}0$ & $T{=}1$ \\
\midrule
Erase & $2.66\!\times$ & $2.48\!\times$ \\
\rowcolor{ourtint}
ALR + IRA & $2.79\!\times$ & $2.58\!\times$ \\
\midrule
Gain over erase (\%) & $4.62\ [3.80,\,5.42]$ & $3.89\ [2.52,\,5.28]$ \\
\bottomrule
\end{tabular}
\end{table}

\paragraph{Label controls.}
Table~\ref{analysis:label-controls} uses the same text recipe with a separate matched pair of training replicates. Gates use probabilities along the selected continuation.

\section{Evaluation Determinism}
\label{app:determinism}

\paragraph{Device consistency.}
The main 8B DFlash-family acceptance-length comparisons and per-prompt analyses use RTX A6000 evaluations; the 4B comparison uses H100, as does the separate EAGLE-3 comparison for both drafter families. An evaluation is bitwise repeatable for the same checkpoint and GPU model, including across hosts, but bf16 near-ties can change the greedy verification path across GPU models. Evaluating identical 8B checkpoints on A100 and RTX A6000 gives relative MAT shifts from $-1.66\%$ to $+0.77\%$ (mean $-0.21\%$, sd 0.46\%), which informs the comparison band below.

\paragraph{Replicate variance.}
The comparison rule uses a pooled seed standard deviation of 0.35\%, based on the estimate of 0.349\% from four RTX A6000 settings. Across all settings evaluated on RTX A6000, the pooled standard deviation is 0.355\%, and the standard deviation across the two data orders is 0.16\%.

\paragraph{Comparison rule.}
We compare relative MAT differences between DFlash-family vision-language methods using a prespecified rule. A tolerance of 1.66\% corresponds to the largest observed cross-device MAT shift with identical checkpoints. The lower and upper thresholds lie two standard deviations of the difference below and above this tolerance. They are 0.96\% and 2.36\% with two training replicates for each method, and 1.09\% and 2.23\% with three. An improvement is supported above the upper threshold. Differences between the thresholds are inconclusive, and those below the lower threshold do not meet the improvement criterion. For non-inferiority of a cheaper setting, we reflect the thresholds below zero. With three training replicates for each method, a difference above $-1.09\%$ meets the non-inferiority criterion, one below $-2.23\%$ does not, and intermediate values remain inconclusive.

\paragraph{Prompt resampling.}
Tables~\ref{results:tab-mat} and~\ref{analysis:schedules} provide the drafter results and erase references used in our prompt-resampling analysis. We resampled the 99 scored prompts of each domain with replacement 10,000 times, applying each resample to both arms and every seed. The 95\% intervals for ALR + IRA's relative gain over erase are 7.71--9.36\% on ALLaVA, 4.39--5.79\% on ShareGPT4V and 6.36--7.91\% with the smaller target. All three intervals remain above zero.

\section{Training Schedules and Rollout Depths}
\label{app:schedules}

Table~\ref{app:epoch-schedules} reports the epoch and envelope sweep, grouping methods by training corpus and epoch count. Table~\ref{app:rollout-depths} holds the epoch count fixed within each group to isolate ALR's rollout depth. Table~\ref{analysis:rollout-depth-summary} summarizes its three-epoch aggregate results. These sweeps select the erase references and the short-rollout ALR setting in Table~\ref{analysis:schedules}. The three-epoch results in Table~\ref{results:tab-mat} instead keep $R=14$ for both ALR and ALR + IRA. ALLaVA times use a standalone reference, with the short-rollout ALR time calibrated against a rollout-depth control measured under the same load. ShareGPT4V ratios use numerator and denominator measurements from runs with the same node-sharing conditions.

Table~\ref{analysis:schedules} reports the one-epoch comparison and the best erase references. The best erase schedule uses three epochs on ALLaVA and one on ShareGPT4V, although differences among erase schedules remain below the preregistered resolution threshold. Figure~\ref{analysis:fig-cost} extends the ALLaVA comparison across training epochs.

\begin{table}[t]
\centering
\caption{Mean acceptance length $\tau$ under different training durations at $T{=}0$. Rollouts use $R=14$. All rows use $\gamma=2$ unless stated otherwise. Subscripts show training-seed standard deviations, and Avg. is the geometric mean across domains.}
\label{app:epoch-schedules}
\small
\setlength{\tabcolsep}{2pt}
\begin{tabularx}{\linewidth}{l*{5}{>{\centering\arraybackslash}X}}
\toprule
Method & Epochs & COCO & TextVQA & DocVQA & Avg. \\
\midrule
\grouprow{6}{ALLaVA} \\
Erase & 1 & $3.106_{\pm0.014}$ & $3.381_{\pm0.028}$ & $4.197_{\pm0.036}$ & $3.533_{\pm0.006}$ \\
ALR & 1 & $3.199_{\pm0.006}$ & $3.593_{\pm0.027}$ & $4.255_{\pm0.020}$ & $3.657_{\pm0.017}$ \\
\rowcolor{ourtint}
ALR + IRA & 1 & $3.263_{\pm0.011}$ & $3.732_{\pm0.006}$ & $4.254_{\pm0.030}$ & $3.728_{\pm0.005}$ \\
\addlinespace[2pt]
Erase & 3 & $3.124_{\pm0.006}$ & $3.375_{\pm0.005}$ & $4.204_{\pm0.037}$ & $3.539_{\pm0.010}$ \\
Erase ($\gamma{=}0.5$) & 3 & $3.090_{\pm0.009}$ & $3.341_{\pm0.004}$ & $4.115_{\pm0.016}$ & $3.489_{\pm0.006}$ \\
ALR & 3 & $3.242_{\pm0.007}$ & $3.655_{\pm0.024}$ & $4.261_{\pm0.016}$ & $3.696_{\pm0.005}$ \\
\rowcolor{ourtint}
ALR + IRA & 3 & $3.376_{\pm0.007}$ & $3.851_{\pm0.011}$ & $4.361_{\pm0.065}$ & $3.841_{\pm0.021}$ \\
\addlinespace[2pt]
Erase & 6 & $3.083_{\pm0.010}$ & $3.283_{\pm0.003}$ & $4.134_{\pm0.034}$ & $3.472_{\pm0.014}$ \\
\midrule
\grouprow{6}{ShareGPT4V} \\
Erase & 0.5 & $3.302_{\pm0.009}$ & $3.408_{\pm0.014}$ & $4.151_{\pm0.029}$ & $3.601_{\pm0.017}$ \\
\addlinespace[2pt]
Erase & 1 & $3.362_{\pm0.018}$ & $3.409_{\pm0.004}$ & $4.208_{\pm0.013}$ & $3.640_{\pm0.002}$ \\
\rowcolor{ourtint}
ALR + IRA & 1 & $3.658_{\pm0.012}$ & $3.542_{\pm0.032}$ & $4.163_{\pm0.014}$ & $3.778_{\pm0.003}$ \\
\addlinespace[2pt]
Erase & 3 & $3.403_{\pm0.010}$ & $3.326_{\pm0.023}$ & $4.023_{\pm0.036}$ & $3.571_{\pm0.016}$ \\
ALR & 3 & $3.614_{\pm0.003}$ & $3.398_{\pm0.009}$ & $4.035_{\pm0.017}$ & $3.673_{\pm0.007}$ \\
\rowcolor{ourtint}
ALR + IRA & 3 & $3.817_{\pm0.024}$ & $3.600_{\pm0.024}$ & $4.073_{\pm0.040}$ & $3.825_{\pm0.029}$ \\
\bottomrule
\end{tabularx}
\end{table}

\begin{table}[t]
\centering
\caption{Mean acceptance length $\tau$ for ALR at different rollout depths $R$, with $T{=}0$. Subscripts show training-seed standard deviations, and Avg. is the geometric mean across domains.}
\label{app:rollout-depths}
\small
\setlength{\tabcolsep}{2pt}
\begin{tabularx}{\linewidth}{l*{4}{>{\centering\arraybackslash}X}}
\toprule
Method & COCO & TextVQA & DocVQA & Avg. \\
\midrule
\grouprow{5}{ALLaVA, one epoch} \\
ALR &  &  &  &  \\
\quad $R{=}14$ & $3.199_{\pm0.006}$ & $3.593_{\pm0.027}$ & $4.255_{\pm0.020}$ & $3.657_{\pm0.017}$ \\
\quad $R{=}6$ & $3.194_{\pm0.009}$ & $3.587_{\pm0.022}$ & $4.250_{\pm0.019}$ & $3.652_{\pm0.001}$ \\
\quad $R{=}4$ & $3.195_{\pm0.006}$ & $3.591_{\pm0.028}$ & $4.236_{\pm0.018}$ & $3.649_{\pm0.014}$ \\
\midrule
\grouprow{5}{ALLaVA, three epochs} \\
ALR &  &  &  &  \\
\quad $R{=}14$ & $3.242_{\pm0.007}$ & $3.655_{\pm0.024}$ & $4.261_{\pm0.016}$ & $3.696_{\pm0.005}$ \\
\quad $R{=}6$ & $3.230_{\pm0.010}$ & $3.629_{\pm0.018}$ & $4.258_{\pm0.008}$ & $3.682_{\pm0.004}$ \\
\quad $R{=}4$ & $3.232_{\pm0.016}$ & $3.606_{\pm0.039}$ & $4.211_{\pm0.063}$ & $3.661_{\pm0.023}$ \\
\quad $R{=}2$ & $3.205_{\pm0.002}$ & $3.489_{\pm0.026}$ & $4.161_{\pm0.032}$ & $3.597_{\pm0.017}$ \\
\midrule
\grouprow{5}{ShareGPT4V, one epoch} \\
ALR &  &  &  &  \\
\quad $R{=}4$ & $3.524_{\pm0.005}$ & $3.460_{\pm0.020}$ & $4.139_{\pm0.019}$ & $3.695_{\pm0.011}$ \\
\midrule
\grouprow{5}{ShareGPT4V, three epochs} \\
ALR &  &  &  &  \\
\quad $R{=}14$ & $3.614_{\pm0.003}$ & $3.398_{\pm0.009}$ & $4.035_{\pm0.017}$ & $3.673_{\pm0.007}$ \\
\quad $R{=}4$ & $3.606_{\pm0.005}$ & $3.384_{\pm0.019}$ & $4.016_{\pm0.019}$ & $3.659_{\pm0.014}$ \\
\bottomrule
\end{tabularx}
\end{table}

\section{Replicates and Robustness}
\label{app:robust}

\paragraph{Training seeds.}
\label{app:seeds}
For the primary vision-language target, the DFlash-family ALLaVA settings use three training replicates, except for erase at six epochs, erase with $\gamma=0.5$ in Table~\ref{app:epoch-schedules} and both regeneration references in Table~\ref{analysis:regeneration}, which use two. Every ShareGPT4V setting, smaller-target setting and text-only method also uses two replicates. Within each setting, the seeds vary training randomness, including anchor sampling, from a common initialization. Corpus selection and data order remain fixed across these replicates. Table~\ref{results:tab-mat} reports means. Tables~\ref{analysis:schedules}, \ref{analysis:rollout-depth-summary}, \ref{app:vision-domains}, \ref{results:text-benchmarks}, \ref{app:epoch-schedules} and~\ref{app:rollout-depths} also give sample standard deviations of acceptance length.

\paragraph{Common-seed check.}
Table~\ref{app:seed-sensitivity} shows that restricting the ALLaVA comparisons to their two common training replicates changes ALR + IRA's gain over the highest measured erase schedule by about 0.1 percentage points at $T{=}0$. The depth comparison still identifies $R=4$ as sufficient and $R=2$ as insufficient, so neither conclusion depends on unequal replicate counts.

\begin{table}[t]
\centering
\caption{Relative MAT differences on ALLaVA at $T{=}0$ using all available training replicates or two matched replicates. Erase uses its highest measured schedule. Rollout depths are compared after three epochs.}
\label{app:seed-sensitivity}
\small
\begin{tabular*}{\linewidth}{@{\extracolsep{\fill}}lcc@{}}
\toprule
Comparison & All seeds (\%) & Common seeds (\%) \\
\midrule
ALR + IRA, 1 epoch vs. erase & $+5.33$ & $+5.24$ \\
ALR + IRA, 3 epochs vs. erase & $+8.54$ & $+8.64$ \\
ALR, $R{=}6$ vs. $R{=}14$ & $-0.37$ & $-0.31$ \\
ALR, $R{=}4$ vs. $R{=}14$ & $-0.93$ & $-0.81$ \\
ALR, $R{=}2$ vs. $R{=}14$ & $-2.68$ & $-2.41$ \\
\bottomrule
\end{tabular*}
\end{table}

\paragraph{Data order, sampling and decoder.}
Figure~\ref{results:fig} compares greedy and sampled decoding using the same methods and benchmarks.

Table~\ref{app:tab-robust} shows that ALR's gain in accepted length over erase persists across data orders on ALLaVA and that sampling at $T{=}1$ preserves the ordering of erase, ALR and ALR + IRA. ALR + IRA remains 5.90\% above erase when each block is verified as a single chain, showing that its improvement does not require tree branching. The mean gain across data orders and the gains under sampling and chain verification exceed the comparison thresholds in Appendix~\ref{app:determinism}. ALR + IRA also outperforms the slot-matched control using either all available replicates or the matched subset in Figure~\ref{analysis:fig-ira}a.

\begin{table}[ht]
\centering
\caption{Relative MAT difference $\Delta$ on ALLaVA from erase outside the main data order, under sampling and with chain decoding, and from a slot-matched control. Sampling uses common random numbers for both arms. Blue shading marks ALR + IRA.}
\label{app:tab-robust}
\small
\begin{tabular}{lcc}
\toprule
Setting & Seeds per arm & $\Delta$ MAT (\%) \\
\midrule
\grouprow{3}{ALR, greedy decoding, other data orders} \\
Data-order replicate 1 & 1 & $+3.99$ \\
Data-order replicate 2 & 1 & $+4.37$ \\
Mean of the two orders & 1 per order & $+4.18$ \\
\grouprow{3}{Sampling at $T{=}1$ on the main data order} \\
ALR & 3 & $+3.23$ \\
\rowcolor{ourtint}
ALR + IRA & 3 & \textbf{$+$7.72} \\
\grouprow{3}{Chain decoding of all 15 slots at $T{=}0$} \\
\rowcolor{ourtint}
ALR + IRA & 3 & \textbf{$+$5.90} \\
\grouprow{3}{Against the slot-matched control at $T{=}0$} \\
\rowcolor{ourtint}
ALR + IRA & 3 and 2 & \textbf{$+$3.92} \\
\bottomrule
\end{tabular}
\end{table}

\paragraph{Displaced-block control.}
Figure~\ref{analysis:fig-ira}a compares ALR + IRA with a control that retains corpus anchors for the blocks IRA would move into target rollouts. The control matches IRA's block count, supervised offsets and slot weights, yet its accepted length remains close to ALR. The contrast tests the joint change in context and continuation labels introduced by IRA.

\paragraph{Choice of erase comparator.}
Table~\ref{results:tab-mat} uses the soft survival weight of PARD-2 as the erase comparator. With the same KD loss and training conditions at $\gamma=2$, the hard gate in Eq.~\eqref{method:gate} scores 0.62\% lower, and removing the gate scores 4.85\% lower. These checks make soft-gated distillation a strong reference for evaluating relabelling.

\paragraph{Difficulty stratification.}
Figure~\ref{analysis:fig-ira}b--c stratifies drafting offsets in held-out captions by the acceptance length of the compared methods' common parent, a distillation drafter trained on the other corpus. These captions are disjoint from the main evaluation. At first-token failures, IRA adds about 0.39 accepted tokens at each offset on both corpora, exceeding the gain from relabelling alone.

\paragraph{Regeneration comparison.}
Table~\ref{analysis:regeneration}a first takes the geometric mean of MAT across COCO, TextVQA and DocVQA within each training replicate, then averages across replicates. The position profiles in Table~\ref{analysis:regeneration}b use one checkpoint of each method and group COCO verification rounds by their starting token offset.

\paragraph{Position profile.}
Table~\ref{analysis:regeneration}b shows that IRA's gain over ALR is larger in the later COCO caption bins for drafters trained on ALLaVA. A seed control shows no such rise. Scoring every drafter on one fixed text preserves this pattern: the mean gain of the last three bins exceeds the first by 3.47 percentage points, compared with 3.45 in the original scoring. For drafters trained on ShareGPT4V, IRA's gain is already large in the first bin, so we report this positional pattern for ALLaVA training.

\paragraph{Regeneration gap.}
Table~\ref{analysis:regeneration}b also shows that the truncated regeneration reference's advantage over ALR grows after the first 16 caption tokens and then levels off. The first-bin gap remains below its 3.83\% comparison threshold. This pattern motivates exposing the drafter to recent target-generated context through IRA.

\section{Speed Measurement}
\label{app:speed}

Table~\ref{app:tab-speed} reports timing results for the trained drafters. The speedup of a domain is the ratio of autoregressive to drafter milliseconds per token in the same run, and the geometric mean over the three domains gives the speedup of a row. AR timing excludes the first target token, and drafter timing excludes the first draft construction. Both timers divide elapsed decoding time by the number of returned output tokens.

The additional vision-language timings in Table~\ref{results:tab-mat} use the same paired AR protocol, with concurrent evaluators on separate H100 GPUs. These measurements also include sampled decoding. The smaller-target ALR speedups average two timing runs, one per training seed.

\begin{table}[ht]
\centering
\caption{Decoding speedup over autoregressive decoding on one H100, by domain and as the geometric mean. Runs gives the timing runs averaged, with the standard deviation of their speedups in parentheses, and MAT carries its seed count $n$. Blue shading marks ALR + IRA.}
\label{app:tab-speed}
\small
\setlength{\tabcolsep}{4pt}
\begin{tabularx}{\linewidth}{Xcccccc}
\toprule
 & & \multicolumn{3}{c}{Speedup per domain ($\times$)} & & \\
\cmidrule(lr){3-5}
Row & Speedup ($\times$) & Caption & TextVQA & DocVQA & Runs & MAT ($n$) \\
\midrule
\grouprow{7}{Trained on the fixed ALLaVA corpus} \\
DFlash, iso-data & 2.09 & 1.86 & 1.99 & 2.46 & 1 & 2.878 (3) \\
Erase (PARD-2 soft survival weight) & 2.62 & 2.28 & 2.55 & 3.10 & 3 (0.03) & 3.539 (3) \\
ALR & 2.68 & 2.22 & 2.77 & 3.13 & 1 & 3.696 (3) \\
ALR, rollout depth $R{=}4$, one epoch & 2.77 & 2.34 & 2.71 & 3.37 & 1 & 3.649 (3) \\
ALR + IRA, one epoch & 2.69 & 2.27 & 2.75 & 3.11 & 1 & 3.728 (3) \\
\rowcolor{ourtint}
ALR + IRA & 2.82 & 2.46 & 2.95 & 3.08 & 4 (0.03) & 3.841 (3) \\
\grouprow{7}{Trained on target-regenerated ALLaVA responses} \\
Regeneration, truncated & 2.89 & 2.49 & 3.01 & 3.23 & 1 & 3.833 (2) \\
\grouprow{7}{Trained on the fixed ShareGPT4V corpus} \\
Erase, one epoch & 2.66 & 2.46 & 2.55 & 3.01 & 1 & 3.640 (2) \\
\rowcolor{ourtint}
ALR + IRA & 2.77 & 2.72 & 2.67 & 2.93 & 1 & 3.825 (2) \\
\bottomrule
\end{tabularx}
\end{table}

\paragraph{Repeated timing.}
Table~\ref{app:tab-speed} averages three paired timing runs for ALLaVA erase and four for ALR + IRA. ALR + IRA's averaged advantage over erase is 0.20$\times$, with a 95\% Welch interval of 0.13--0.26$\times$. Repeating ALR + IRA at the start and end of the first session gave a 0.62\% drift, below the prespecified 3\% threshold. Autoregressive timing had a coefficient of variation of 3.4--5.6\% across the three domains in that session, so we base close comparisons on accepted length rather than small speed differences.

\section{H100 Comparison with EAGLE-3}
\label{app:eagle-h100}

Table~\ref{app:eagle-h100-table} compares the ALLaVA drafters with the same Qwen3-VL-8B target on H100. EAGLE-3 (HF) uses the official dynamic-tree proposal \citep{li2025eagle3} in our Hugging Face adapter, with 60 packed nodes including the anchor, depth 7 and top-$K$ 10 fixed before evaluation. Both families use the common target runtime and, at $T{=}1$, target sampling with common random numbers at each output position. Each method uses one checkpoint and one evaluation of each domain, with 100 prompts, one excluded warmup and a 256-token output limit.

Table~\ref{app:eagle-h100-table} uses the paired AR reference and the decoding window defined in Appendix~\ref{app:speed}, excluding the first draft construction. This comparison uses a separate H100 timing session from the main results. Including the first draft construction gives EAGLE-3 speedups of 1.26$\times$ at $T{=}0$ and 1.23$\times$ at $T{=}1$. These H100 acceptance lengths are separate from the A6000 measurements in the main table.

\begin{table}[h]
\centering
\caption{Observed H100 decoding speedup over AR and mean acceptance length $\tau$ for Qwen3-VL-8B drafters trained on ALLaVA. Values are geometric means across COCO, TextVQA and DocVQA. EAGLE-3 (HF) denotes our Hugging Face adapter.}
\label{app:eagle-h100-table}
\small
\begin{tabular}{lcccc}
\toprule
 & \multicolumn{2}{c}{$T{=}0$} & \multicolumn{2}{c}{$T{=}1$} \\
\cmidrule(lr){2-3}\cmidrule(lr){4-5}
Method & Speedup & $\tau$ & Speedup & $\tau$ \\
\midrule
EAGLE-3 (HF) & $1.27\times$ & $2.32$ & $1.24\times$ & $2.29$ \\
DFlash & $2.09\times$ & $2.87$ & $2.04\times$ & $2.78$ \\
Erase & $2.65\times$ & $3.54$ & $2.49\times$ & $3.38$ \\
ALR & $2.68\times$ & $3.69$ & $2.55\times$ & $3.47$ \\
\rowcolor{ourtint}
ALR + IRA & $\mathbf{2.85}\times$ & $\mathbf{3.84}$ & $\mathbf{2.68}\times$ & $\mathbf{3.63}$ \\
\bottomrule
\end{tabular}
\end{table}

\section{Limitation}

Our implementation trains feature-conditioned block drafters and requires access to the target's hidden states and token distributions. Its rollouts are greedy, while the resulting drafter supports both greedy and sampled decoding. Constructing rollouts for the sampling policy used at inference is a natural extension.

\end{document}